\documentclass{article}
\PassOptionsToPackage{numbers, compress}{natbib}
\usepackage[preprint]{neurips_2026}

\usepackage{neurips_2026}
\usepackage[normalem]{ulem}
\usepackage[utf8]{inputenc}
\usepackage[T1]{fontenc} 
\usepackage{hyperref}       
\usepackage{url}            
\usepackage{booktabs}       
\usepackage{amsfonts}       
\usepackage{nicefrac}       
\usepackage{microtype}      
\usepackage{tcolorbox} 
\tcbuselibrary{breakable}
\usepackage{tabularx}
\usepackage{array}
\usepackage{textcomp}
\usepackage{pifont} 
\usepackage{caption}
\usepackage{pifont}
\usepackage[table,xcdraw]{xcolor}  
\usepackage{amsmath}                    
\usepackage{amssymb}                    
\usepackage{bm}                         
\usepackage{graphicx}                  
\usepackage{multirow}                   
\usepackage{float}                      
\usepackage{enumitem}              
\usepackage[font=small,labelfont=bf]{caption}   

\usepackage{listings}
\lstdefinestyle{jsonTiny}{
  basicstyle=\ttfamily\tiny,
  breaklines=true,
  breakatwhitespace=false,
  columns=fullflexible,
  keepspaces=true,
  showstringspaces=false,
  frame=none
}
\usepackage{subcaption}   
\usepackage{xspace}
\definecolor{lightblue}{RGB}{229,242,255}

\newcommand{\benchmark}{TRACE\xspace}

\newcommand{\modulefull}{Feature Recovery Module\xspace}
\newcommand{\module}{FRM\xspace}

\newcommand{\boldheader}[1]{\textbf{#1}}

\NewDocumentCommand{\heng}
{ mO{} }{\textcolor{red}{\textsuperscript{\textit{Heng}}\textsf{\textbf{\small[#1]}}}}

\NewDocumentCommand{\daf}
{ mO{} }{\textcolor{red}{\textsuperscript{\textit{DAF}}\textsf{\textbf{\small[#1]}}}}

\NewDocumentCommand{\aditi}
{ mO{} }{\textcolor{blue}{\textsuperscript{\textit{Aditi}}\textsf{\textbf{\small[#1]}}}}

\NewDocumentCommand{\savya}
{ mO{} }{\textcolor{orange}{\textsuperscript{\textit{Savya}}\textsf{\textbf{\small[#1]}}}}

\NewDocumentCommand{\sofia}
{ mO{} }{\textcolor{purple}{\textsuperscript{\textit{Sofia}}\textsf{\textbf{\small[#1]}}}}

\NewDocumentCommand{\zhenhailong}
{ mO{} }{\textcolor{blue}{\textsuperscript{\textit{Zhenhailong}}\textsf{\textbf{\small[#1]}}}}

\NewDocumentCommand{\jeongh}
{ mO{} }{\textcolor{brown}{\textsuperscript{\textit{Jeonghwan}}\textsf{\textbf{\small[#1]}}}}

\renewcommand{\cite}{\citep} 

\definecolor{hlcolor}{RGB}{255, 243, 205} 

\title{Formatting Instructions For NeurIPS 2026}

\author{%
  Aditi Tiwari\thanks{Corresponding author: \texttt{aditit5@illinois.edu}} \quad
  Sofia Stoica \quad
  Savya Khosla \quad
  David Forsyth \quad
  Heng Ji \\
  University of Illinois Urbana-Champaign \\
  Urbana, IL 61801 \\
  \texttt{\{aditit5, sstoica2, savyak2, daf, hengji\}@illinois.edu}
}

\title{Feature Recovery for Object Understanding After Irreversible Fire Damage}

\begin{document}

\maketitle

\begin{abstract}
Objects in post-fire environments often undergo irreversible physical transformations that change their geometry, material state, and visual appearance. Detecting and identifying these remnants is critical for locating hazards, reconstructing pre-incident contents, and inventorying losses. Unlike standard image corruptions, these degradations affect the physical structure of the object itself. To study this setting, we introduce \textbf{\benchmark}, a transformation-aware benchmark for post-fire object understanding. \benchmark contains 21.4K real-image-grounded synthetic scenes and paired object-level pristine-to-degraded progressions spanning 499 object identities across 189 categories. We define five tasks targeting localization and pre-degradation understanding: degraded-object detection, pristine-state recovery and retrieval, original material recovery, pristine description generation, and functional reasoning. Existing models degrade sharply with severity. From the least to the most severe level, RF-DETR mAP decreases by 71\% relative, while InternVL3.5 retrieval R@1 falls from 93.85 to 28.11. To address this, we propose \textbf{\modulefull} (\module), a lightweight plug-and-play module that maps degraded encoder features to pristine-aligned representations while keeping the host frozen. Trained only with paired feature supervision, \module improves scene-level detection, CLIP/SigLIP2 feature recovery, and all four object-level VLM tasks, with larger gains under more severe degradation. Across VLM hosts and severity levels, relative gains average 12.5\% for retrieval, 20.1\% for material recovery, 13.2\% for description generation, and 12.4\% for functional reasoning.
\end{abstract}


\vspace{-1em}
\section{Introduction}
\label{sec:intro}
\vspace{-0.5em}
Real-world post-disaster environments present a fundamental challenge for visual perception. Unlike classical image corruptions, post-fire degradation is physical and irreversible. Wood chars, plastic warps, metal oxidizes, and glass fractures, jointly altering geometry, material state, and surface appearance. Recovering object identity, material, and function from such remnants is critical across post-disaster workflows, including firefighters locating hazardous items, investigators reconstructing pre-incident contents, and adjusters inventorying damaged property from partial remains.

\begin{figure}[t]
    \centering
    \includegraphics[width=\linewidth]{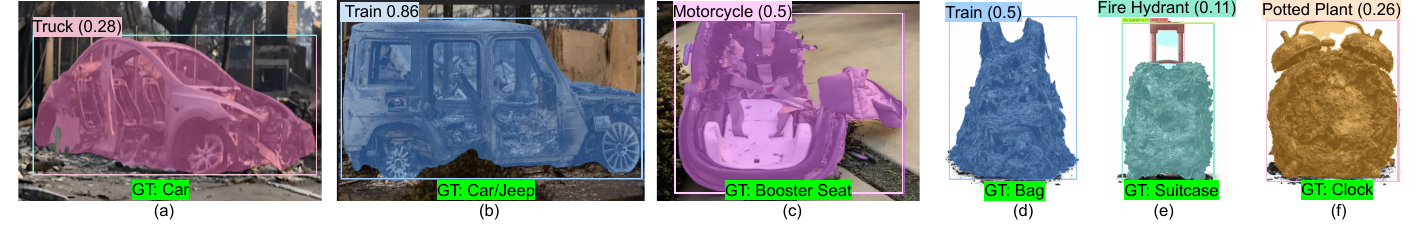}
    \caption{\textbf{RF-DETR failures under post-fire degradation.} RF-DETR consistently mislabels detected objects in real post-fire scenes (a-c) and synthetic degradations (d-f), as post-fire transformations alter object structure, material, and surface appearance in ways not seen by the model. This highlights the need for benchmarks that explicitly link pristine objects to their degraded counterparts, which are not available in current datasets.}
    \label{fig:rfdetr_failures}
    \vspace{-1.2em}
\end{figure}

Existing benchmarks do not capture this setting. Standard object detection benchmarks~\cite{lin2014microsoft, Everingham2010, xiaosun2010, russakovsky2015imagenet} contain mostly pristine object instances, while robustness suites~\cite{schmalfuss2025robustspring, Exdark} perturb appearance over a fixed geometric scaffold, leaving object shape and material identity largely intact. In contrast, post-fire transformation changes structure, material, and appearance together. Across RF-DETR~\cite{rf-detr}, CLIP~\cite{radford2021learning}, SigLIP2~\cite{siglip2}, and five VLMs, Qwen3-VL~\cite{bai2025qwen3}, Qwen2.5-VL~\cite{qwen2.5}, InternVL3.5~\cite{wang2025internvl3}, Molmo2~\cite{clark2026molmo2}, and LLaVA-OV-1.5~\cite{an2025llava}, we observe consistent failure. From the least (L0) to the most severe degradation (L4) level, RF-DETR mAP drops by 71\%, and 92.9\% of boxes localized at IoU$\geq$0.5 are misclassified (Figure~\ref{fig:rfdetr_failures}). VLMs, despite access to world knowledge, exhibit the same degradation pattern: InternVL3.5 retrieval R@1 falls from 93.85 to 28.11, and Qwen2.5-VL falls from 76.92 to 32.31. Similar drift in frozen encoder features indicates a systematic limitation of current visual representations rather than a model-specific deficiency.

\begin{figure}[ht]
    \centering
    \includegraphics[width=\linewidth, height=0.5\linewidth]{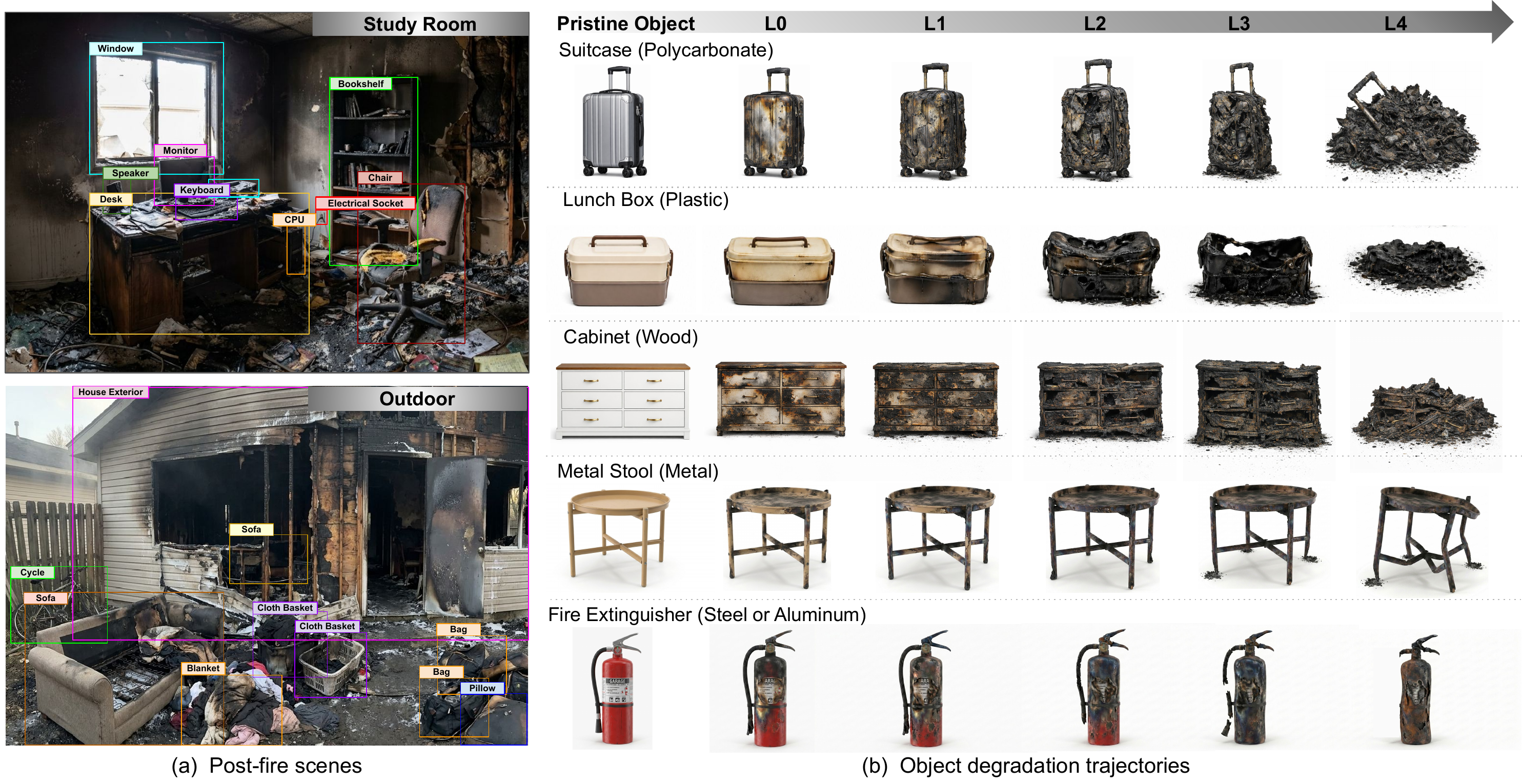}
    \vspace{-1.5em}
\caption{\textbf{\benchmark dataset overview.} \benchmark contains real-image-grounded synthetic post-fire scenes and paired pristine-to-degraded object trajectories. Scene annotations include bounding boxes, material labels, and captions; object trajectories include part-level materials and progressive degradation states for tracking identity changes under physical transformation. Detailed object and scene statistics are provided in Appendix~\ref{app:dataset-details}.}
    \label{fig:degradation_levels}
    \vspace{-1.2em}
\end{figure}

To study this gap, we introduce \textbf{\benchmark}, a transformation-aware benchmark for post-fire object understanding. \benchmark provides 21.4K real-image-grounded synthetic \textbf{post-fire scenes} with bounding boxes, material labels, and degradation-severity captions (Figure~\ref{fig:degradation_levels}a). It also provides \textbf{object degradation trajectories} for 499 instances across 189 object types, each with a pristine reference and five progressively degraded states annotated at the part level (Figure~\ref{fig:degradation_levels}b). On \benchmark, we define five tasks spanning localization and pre-degradation understanding (Table~\ref{tab:task-overview}). These are degraded-object detection, pristine-state recovery and retrieval, original material recovery, pristine description generation, and functional reasoning.

Pixel-space restoration is a natural baseline, but physical degradation makes pristine reconstruction ambiguous. Irreversible changes can destroy object parts and material cues that cannot be uniquely inferred. We fine-tune Restormer~\cite{Zamir2021Restormer} on \benchmark degraded-pristine pairs and use it before frozen RF-DETR and VLM retrieval, but even with paired supervision, Restormer gives only modest gains and remains below feature recovery. We instead train a lightweight residual \textbf{\modulefull (\module)} at the host feature interface, leaving all host weights frozen. \module maps degraded features toward pristine counterparts in the host representation space, leveraging semantics already encoded by pretrained models. Across detectors, encoders, and VLMs, \module yields severity-scaled gains: RF-DETR detection improves by 30.8\% relative, while InternVL3.5 and Qwen3-VL improve by 27\% and 12\% on average across the four VLM pristine-state tasks.

\noindent\textbf{Contributions.}
\begin{itemize}[leftmargin=*, itemsep=2pt, topsep=2pt]
    \item \textbf{Benchmark.} \benchmark provides 21.4K real-image-grounded synthetic post-fire scenes with object-level annotations and 499 paired pristine-to-degraded object trajectories across 189 types, supporting five evaluation tasks (Table~\ref{tab:task-overview}).
    \vspace{-0.6em}
    \item \textbf{Findings.} On \benchmark, state-of-the-art detectors, encoders, and VLMs fail consistently under physical degradation, indicating a representational rather than model-specific limitation.
      \vspace{-0.6em}
    \item \textbf{Method and release.} \module is a plug-and-play residual feature-recovery module that improves all five evaluation tasks in Table~\ref{tab:task-overview} across diverse hosts. RF-DETR detection improves by $30.8\%$ relative, and the four VLM pristine-state tasks improve by 12.4 \% to 20.1\% mean relative across five frozen VLMs. We release the dataset, generation and evaluation pipelines, and training code.
\vspace{-0.5em}
\end{itemize}

\begin{table}[t]
    \centering
    \tiny
    \setlength{\tabcolsep}{5pt}
    \renewcommand{\arraystretch}{1}
    \caption{\textbf{Evaluation tasks in post-fire object understanding.} We evaluate five tasks spanning detection, retrieval, and reasoning, with pristine-state recovery split into feature-space and gallery-retrieval protocols. Full prompts are provided in Appendix~\ref{app:task-details}.}
    \label{tab:task-overview}
    \rowcolors{2}{gray!7}{white}
    \begin{tabular}{p{0.20\linewidth}p{0.29\linewidth}p{0.24\linewidth}p{0.20\linewidth}}
    \toprule
    \rowcolor{gray!22}
    \textbf{Task} & \textbf{Goal} & \textbf{Output} & \textbf{Metric} \\
    \midrule
Degraded-object detection
    & Localize all objects of interest in scene-level post-fire images
    & Bounding boxes + mapped COCO classes
    & COCO-mapped mAP@0.5:0.95 \\

Pristine-state recovery (feature)
    & Recover CLIP/SigLIP2 features toward paired pristine targets
    & Recovered visual embeddings
    & MSE, cosine similarity \\

Pristine-state retrieval (gallery)
    & Retrieve paired pristine reference from candidate gallery using VLM embeddings
    & Ranked pristine gallery
    & Recall@1 \\

Original material recovery
    & Predict pre-degradation materials from degraded object image
    & Material label set
    & Micro-F1 over normalized material classes \\

Pristine description generation
    & Describe object identity, material, and function before degradation
    & Single-sentence description
    & Mean of identity accuracy, material micro-F1, and function accuracy \\

Functional reasoning
    & Predict pre-degradation function and required restoration materials
    & Function tags + restoration materials
    & Mean of function-tag micro-F1 and restoration-material micro-F1 \\
    \bottomrule
    \end{tabular}
    \vspace{-2.4em}
\end{table}


\vspace{-1em}
\section{Related Work} 
\vspace{-0.5em}

Prior work on visual robustness studies models under image corruptions, domain shift, adverse conditions, synthetic distribution shifts, or pixel-level restoration. These settings assume the underlying object is intact and recoverable from the observation.
\textbf{i. Benchmarks.} Robustness benchmarks~\cite{hendrycks2019benchmarking, barbu2019objectnet, sakaridis2021acdc, Arsenos_flight, chong2025towards, michaelis2019dragon, hao2026clearunlockinggenerativepotential, schmalfuss2025robustspring, Exdark} evaluate recognition under appearance shifts (noise, blur, weather, viewpoint) but preserve object structure and material identity. Natural-disaster benchmarks~\cite{gupta2019xbd, rahnemoonfar2023rescuenet, rahnemoonfar2021floodnet, weber2022incidents1m, proma2022nadbenchmarks} target damage assessment from aerial or social-media imagery at the building or scene level, not at the level of individual objects undergoing irreversible transformation. \benchmark is the first benchmark to study object-level physical degradation with paired pristine-to-degraded trajectories, supporting severity-stratified evaluation and feature-space recovery.
\textbf{ii. Methods.} The methods most relevant to \module fall into three families, each with a different assumption about the nature of degradation.
\textit{Domain adaptation} aligns features between source and target distributions through covariance matching, pseudo-labeling, or prototype alignment~\cite{Sun2016DeepCC, wang2021exploring, chen2025datr, lavoie2025large, yan2026compositional, ramamonjison2021simrod}. These methods assume a coherent target distribution to align to; under our setting, each degradation severity induces a different distribution, and the relevant supervision is paired pristine-degraded correspondences rather than aggregate distributional alignment.
\textit{Test-time adaptation} adapts models on unlabeled test data through pseudo-supervision, feature-distribution alignment, or dynamic class statistics~\cite{ruan2024Iou, liu2024mlfa, zhou2025bayesian}. These methods optimize at deployment time without paired pristine references; \module instead trains once on paired data and applies without per-deployment optimization.
\textit{Image restoration} recovers high-quality images from degraded image quality observations~\cite{Zamir2021MPRNet, liang2021swinir, Zamir2021Restormer, chu2022nafssr, luo2023controlling, deng2025learning, ren2026efficient, yan2026compositional}, and feature-quality methods characterize degradation-induced shifts in learned embeddings~\cite{BIANCO2021128, agnolucci2024arniqa, becker2026self}. Restoration shares \module’s motivation but operates in pixel space, where physical degradation makes inversion ill-posed and errors can propagate to downstream recognizers. Feature-quality methods quantify degradation without correcting it, whereas \module uses paired supervision to recover pristine-aligned features in the host representation space while keeping all host weights frozen.

\vspace{-1em}
\section{\benchmark}
\label{sec:benchmark}
\vspace{-0.5em}

\boldheader{Design Principles. }\benchmark is shaped by three constraints specific to physical transformations of the object.
\textit{i. Pairing.} Each object appears in both pristine and degraded states with instance-level correspondence. This enables direct measurement of representation drift under transformation, rather than end-task accuracy under shifted distributions alone.
\textit{ii. Severity progression.} Each object is rendered along a discrete five-level severity ladder (L0-L4), where L0 denotes mild surface damage and L4 denotes severe deformation or near-total destruction. This enables severity-stratified evaluation and exposes whether a method interpolates within a trained severity range or extrapolates beyond it (Sec.~\ref{sec:experiments}).
\textit{iii. Real-world grounding.} Synthetic generation is conditioned on real post-fire imagery~\cite{tiwari2025fire360}, grounding visual statistics, material behavior, and scene composition in observed environments rather than free-form prompts alone. Each sample includes object class, material composition, degradation-state descriptions, failure modes, embeddings, and bounding boxes; see Appendix~\ref{app:dataset-details}. To verify annotation reliability, two authors independently assigned accept/reject labels on a stratified 300-object subset, yielding 93.4\% agreement and Cohen's $\kappa=0.836$.

\boldheader{Object Inventory.}
The object inventory targets two post-fire response needs. The first covers hazardous items such as aerosol cans, gas cylinders, and chemical containers that responders must identify and secure. The second covers personal valuables and everyday objects, including electronics, furniture, bags, and medical equipment, that matter for insurance assessment and inventory reconstruction. Overall, \benchmark spans 189 object types and 11 normalized material classes, with plastic (1{,}315 mentions), metal (948), fabric/textile (390), and wood (287) most represented.

\boldheader{Scene-Level Subset.}
Gemini~3 Pro Image~\cite{google2025gemini3proimage} generates each scene from a structured prompt and two to four real post-fire references, sampled from approximately 700 images drawn primarily from Fire360~\cite{tiwari2025fire360} and research-permissive web sources. References span residential, commercial, institutional, and outdoor environments. Scene prompts cover four environment groups and 30 scene types. A Gemini~3 Flash \cite{google2025gemini3flash} vision-language judge scores each candidate for realism, material fidelity, structural plausibility, and consistency with the references. Candidates below 0.7 are rejected, yielding 21{,}400 accepted scenes from approximately 29{,}000 generations. Full prompts, judge criteria, filtering details, and licensing notes are in Appendix~\ref{app:dataset}.

\boldheader{Object-Level Subset.}
Each object trajectory contains one pristine render and five progressively degraded states (L0-L4) generated with material-aware prompts that capture wood charring, plastic melting, metal oxidation, and glass fracture. Trajectories span 499 distinct object identities across 189 object types, yielding 2{,}994 object-level images. Some object types include multiple subtypes or variants, which are treated as distinct identities when constructing train, validation, and test splits. The object-level subset is split 80/10/10 by object identity, ensuring that all renders of an identity remain in the same split.

\boldheader{Real Post-Fire Crops. }
To evaluate beyond synthetic data, we manually crop 500 objects from the pool of approximately 700 real post-fire reference images described above. Each crop is annotated with identity and severity labels following the same protocol as the synthetic subset. These real crops are included in the object-level test set.

\boldheader{Evaluation Tasks.} \benchmark defines five tasks spanning detection, retrieval, and reasoning in post-fire settings (Table~\ref{tab:task-overview}). \textbf{(1) Degraded-object detection.} First responders must rapidly locate hazardous or structurally compromised objects in post-fire scenes. Given a scene-level image, the task is to detect and localize all objects of interest, evaluated with class-aware COCO-mapped mAP@0.5:0.95 at each degradation level. Only boxes with unambiguous COCO mappings~\cite{lin2014microsoft} are included in mAP. The mapping covers 55 of the 80 COCO categories. \textbf{(2) Pristine-state recovery and retrieval.} Matching a damaged remnant to its undamaged counterpart is the foundation for downstream recovery. This retrieval setting is similar to TOR~\cite{tiwari2025fire360}, but TOR defines transformed-object retrieval in video space, while \benchmark evaluates paired pristine-to-degraded recovery at the object level. We evaluate this in two protocols. In feature space, CLIP and SigLIP2 features are recovered toward paired pristine targets and measured by MSE and cosine similarity. In gallery retrieval, VLM visual embeddings retrieve the paired pristine reference from a candidate gallery, measured by Recall@1. \textbf{(3) Material recovery.} Knowing original materials determines whether an object is restorable or requires replacement and helps environmental teams triage debris for safe disposal. Given a degraded object image, the model predicts the object's pre-degradation materials. Outputs are mapped to our normalized material vocabulary and evaluated using micro-F1 over the 11 material classes. \textbf{(4) Pristine description generation.}
Reconstruction planning and insurance documentation require a concise account of what the object was, its material composition, and its intended use. The model generates a one-sentence description of the object's pre-degradation identity, materials, and function. We parse the description into identity, material, and function fields. Identity and function are scored by categorical accuracy after synonym normalization, with GPT5 \cite{singh2025openai} as LLM-judge fallback only for unresolved paraphrases. Materials are scored with micro-F1 over the normalized material vocabulary. The final description score is the unweighted mean of identity accuracy, material micro-F1, and function accuracy, reported on a 0-100 scale. \textbf{(5) Functional reasoning.} Fire investigators and restoration teams need structured judgments beyond description, such as whether an object could have served as an ignition source, fuel, or hazard, and what materials are needed for repair. The model predicts pre-degradation function tags and restoration materials, evaluated as the unweighted mean of micro-F1 over function tags and micro-F1 over restoration-material labels. Appendix~\ref{app:task-details} specifies the prompts, output normalization, label vocabularies, and exact score computations for all generation-based tasks.

\begin{figure}[t]
    \centering
    \includegraphics[width=\linewidth, height=0.4\linewidth]{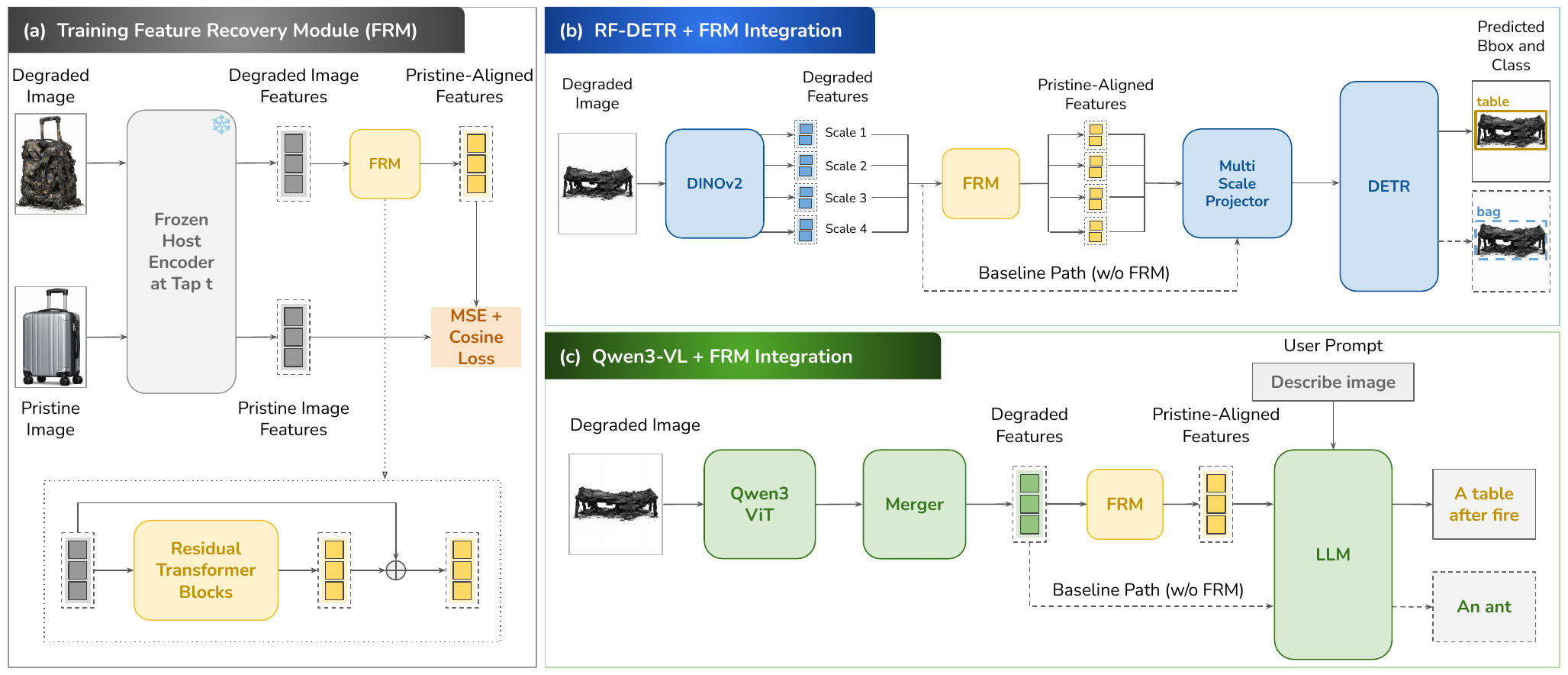}
    \caption{\textbf{\modulefull (\module).} (a) \module is a residual Transformer~\cite{transformer} over visual tokens, trained to recover pristine-aligned features from degraded ones with a zero-gated identity initialization; tap $t$ is the frozen host feature interface where \module is inserted. (b) In RF-DETR, DINOv2~\cite{oquab2023dinov2} feature maps from four tapped layers are flattened, corrected by a \emph{single shared} \module head, and reshaped before the unchanged detector. (c) In Qwen3-VL, \module corrects post-merger visual tokens before they are consumed by the frozen language model. Dashed paths denote the no-\module baseline.}
    \label{fig:frm}
    \vspace{-1.2em}
\end{figure}

\vspace{-1em}
\section{\modulefull (\module)}
\label{sec:frm}
\vspace{-0.5em}
\benchmark provides paired observations of each object instance before and after physical transformation (Figure~\ref{fig:frm}). This pairing enables direct supervision of feature-space recovery: for any frozen visual encoder, the degraded image serves as input and the paired pristine image defines the target representation. We operate in feature space rather than pixel space because severe physical transformations alter geometry, material state, and surface structure in ways that make pixel-level inversion ambiguous. Since \module modifies only the visual feature representations, the encoder, projector or merger, detector head, and language model remain frozen.

\boldheader{Feature-Recovery Problem. }
Let $x_i^{p}$ denote the pristine image of object instance $i$, and let $x_i^{\ell}$ denote a paired image of the same instance after transformation at degradation level $\ell\in\{0,\dots,4\}$. For a frozen encoder $E$ and a tap point $t$ (an intermediate interface where features can be read and written back to), let $E_t(x)\in\mathbb{R}^{N_t\times D_t}$ denote the features at $t$. \module learns a shape-preserving function $h_\theta:\mathbb{R}^{N_t\times D_t}\to\mathbb{R}^{N_t\times D_t}$ that maps degraded features toward their pristine counterparts:
\[
\tilde{z}_i^{\ell,t}\;=\;h_\theta\!\big(E_t(x_i^{\ell})\big)\;\approx\;E_t(x_i^{p}).
\]The corrected features $\tilde{z}_i^{\ell,t}$ are forwarded through the unchanged downstream pipeline. As detailed below, $h_\theta$ is implemented as a residual correction $h_\theta(z) = z + \Delta_\theta(z)$ with $\Delta_\theta=0$ at initialization.
 
\boldheader{Architecture. }
\module is a residual Transformer ($M$ blocks, default $M{=}4$ for
VLMs, $M{=}1$ for RF-DETR) with zero-gated initialization. For any tapped feature tensor $z\in\mathbb{R}^{B\times N\times D}$ passed to \module, let $u^{0}=z$. Block $m\in\{1,\dots,M\}$ applies
\[
\hat{u}^{m} = u^{m-1} + \alpha^{m}_{\mathrm{attn}}\odot\mathrm{MSA}(\mathrm{LN}(u^{m-1})),\quad
u^{m} = \hat{u}^{m} + \alpha^{m}_{\mathrm{mlp}}\odot\mathrm{MLP}(\mathrm{LN}(\hat{u}^{m})),
\]
with $H$ attention heads, an MLP of hidden width $rD$, and learned channel-wise residual scales $\alpha^{m}_{\mathrm{attn}},\alpha^{m}_{\mathrm{mlp}}\in\mathbb{R}^{D}$ initialized to zero. We define the residual correction $\Delta_\theta(z) := u^{M} - z$, so that $h_\theta(z) = z + \Delta_\theta(z)$. Because the residual scales gate all MSA and MLP outputs to zero, $\Delta_\theta(z)=0$ at initialization regardless of inner weights, and \module acts as the identity on the host interface, which stabilizes early training. Unlike encoder fine-tuning, \module never touches encoder weights, preserving the downstream model's training-time feature distribution.
 
\boldheader{Tap-Point Selection. }The tap point $t$ involves a tradeoff between spatial detail and downstream alignment. Earlier taps in the visual encoder preserve higher spatial resolution, making local degradation patterns easier to identify and correct at the patch level. Later taps, and in particular post-projector or post-merger interfaces, produce features in the same space that the downstream head was trained on, so corrections at these locations translate more directly into task performance. We select the tap point for each host based on this tradeoff and ablate the choice in Table~\ref{tab:app-frm-ablations}.

\boldheader{Training Objective. }For a paired sample $(x_i^{\ell},x_i^{p})$, the encoder extracts $z_i^{\ell,t}=E_t(x_i^{\ell})$ and $z_i^{p,t}=E_t(x_i^{p})$ at the same tap. The pristine branch is computed without gradient tracking, and \module is applied only to the degraded branch, $\tilde{z}_i^{\ell,t}=h_\theta(z_i^{\ell,t})$. Dropping indices, the recovery loss combines element-wise MSE with token-level cosine alignment:
\vspace{-0.2em}
\[
\mathcal{L}_{\module}(\tilde{z},z^{p})
= \tfrac{1}{BND}\|\tilde{z}-z^{p}\|_{2}^{2}
\;+\;\lambda_{\cos}\Big(1-\tfrac{1}{BN}\sum_{b,n}\cos(\tilde{z}_{b,n},z^{p}_{b,n})\Big),
\]
\vspace{-0.2em}
with $\lambda_{\cos}=0.1$ selected on the validation split.(Sec.~\ref{sec:exp-ablations}). The MSE term constrains token magnitudes in the downstream feature space, and the cosine term penalizes directional drift. Spatial augmentations are restricted to transformations that preserve patch-grid correspondence (horizontal flips, $90^{\circ}$ rotations, and stride-aligned resizes). The same transformation is sampled once and applied to both $x_i^{\ell}$ and $x_i^{p}$ before encoding. Arbitrary-offset crops are excluded because they break paired token correspondence.
 
\boldheader{Training Cost and Data Extensibility. }Because only \module is trained and the encoder is frozen, encoder features can be precomputed once per dataset and reused across epochs. On a single NVIDIA A100, training an \module head to convergence on approximately \textbf{2,000 paired examples} takes approximately \textbf{one hour} with cached encoder features and approximately \textbf{three hours} when encoder forward passes are recomputed on the fly. Adding new transformation categories or object instances requires only encoding the additional images and continuing \module training.

\boldheader{Instantiation in RF-DETR.} RF-DETR uses a DINOv2 encoder, a multi-scale projector, a DETR transformer, and class/box heads. We insert \module after the encoder and before the projector, and all other components remain frozen. For RF-DETR-Medium with $576\times 576$ inputs and patch size $16$, we extract features from four evenly spaced DINOv2-Small encoder layers $s\in\{3,6,9,12\}$, each producing a $36\times 36$ grid with $z^{s}\in\mathbb{R}^{B\times 384\times 36\times 36}$. Each map is flattened to $u^{s}=\phi(z^{s})\in\mathbb{R}^{B\times 1296\times 384}$ and corrected by a single shared \module head $h_{\theta}$ applied to all four scales:
\[
\tilde{u}^{s} = h_{\theta}(u^{s}),\qquad
\tilde{z}^{s}=\phi^{-1}(\tilde{u}^{s}),\qquad s\in\{3,6,9,12\}.
\]

Sharing $h_\theta$ across scales is possible because all four layers produce the same hidden dimension ($D{=}384$), and keeps the parameter count independent of the number of tapped layers. The recovery loss $\mathcal{L}_{\module}$ is applied to each scale independently and averaged over the four tapped layers.

\boldheader{Instantiation in Qwen3-VL. }Qwen3-VL consists of a frozen vision transformer, a frozen merger that compresses each $2\times2$ spatial neighborhood into one token and projects visual features into the LLM-facing space, and a frozen language model. With $448\times448$ inputs and patch size $16$, the post-merger visual features have shape $z_{\mathrm{post}}\in\mathbb{R}^{B\times196\times2560}$. We insert \module at this post-merger interface, so the corrected tokens $\tilde{z}_{\mathrm{post}}$ are passed directly to the frozen language model. This tap aligns \module with the representation actually consumed by the LLM. Instantiation details for CLIP, SigLIP2, and the remaining VLM hosts follow the same recipe and are provided in Appendix~\ref{app:frm-hosts}.

\boldheader{Plug-and-Play Scaling and Overhead.}
Only the token dimension $D$, capacity $(M,r)$, and tap point are configured per host, where $D$ is determined by the selected tap point. The \module head has approximately $P_{\module}\approx M(4+2r)D^2$ parameters ($4D^2$ from multi-head self-attention, $2rD^2$ from the MLP). Capacity can be selected from an overhead budget $\varepsilon$ via $M(4+2r)\le \varepsilon P_{\mathrm{host}}/D^2$. We instantiate \module with a 1.77M-parameter head for RF-DETR-Medium ($P_{\mathrm{host}}{=}33.7$M, $D{=}384, M{=}1, r{=}4$, 5.25\% overhead) and a 314.6M-parameter head for Qwen3-VL-4B ($P_{\mathrm{host}}{=}4.0$B, $D{=}2560, M{=}4, r{=}4$, 7.86\% overhead).

\boldheader{Inference and Deployment. }
At inference, \module is inserted at its trained tap. The frozen encoder produces degraded features, \module recovers pristine-aligned features, and the corrected features continue through the unchanged downstream pipeline. \module weights are tied to a specific (encoder, tap, feature space) configuration, so each new host requires a separately trained module.
  
\vspace{-0.5em}
\section{Experiments}
\label{sec:experiments}
\vspace{-0.5em}

\boldheader{Experimental setup.}
We evaluate on the scene-level and object-level \benchmark subsets from Sec.~\ref{sec:benchmark}. Degraded-object detection uses the synthetic scene-level test set, while the four object-level tasks use held-out degraded trajectories over L0-L4, with a separate 500-crop real post-fire evaluation in Table~\ref{tab:ablations}. Tasks and metrics follow Table~\ref{tab:task-overview}, with prompts, normalization rules, and exact score computations specified in Appendix~\ref{app:task-details}. For detection, we compare \module against RF-DETR fine-tuning, RF-DETR domain adaptation (SFA~\cite{wang2021exploring}, DATR~\cite{chen2025datr}), and pixel-restoration preprocessors before frozen RF-DETR, including WildIR~\cite{wildir} and Restormer~\cite{Zamir2021Restormer} fine-tuned on \benchmark. We also evaluate pixel restoration before frozen VLMs: Restormer for degraded-to-pristine gallery retrieval and WildIR for open-ended pristine-state object recovery. All trainable baselines use the \benchmark training split; hosts remain frozen unless marked as fine-tuned. For feature-space alignment, we evaluate frozen CLIP ViT-L/14 and SigLIP2 encoders. For VLM tasks, the baseline is the frozen host on degraded inputs, and the pristine ceiling is the same host on pristine inputs. We evaluate five frozen VLM hosts, Qwen3-VL-4B, Qwen2.5-VL-3B, InternVL3.5-4B, Molmo2-4B, and LLaVA-OneVision-1.5-4B, using deterministic decoding and identical prompts, parsers, normalization rules, and scoring scripts with and without \module. All experiments are run on NVIDIA A100 GPUs.

\vspace{-1em}
\subsection{Degraded-Object Detection and Feature-Space Recovery}
\label{sec:exp-rep}
\vspace{-0.5em}
\vspace{-0.45em}

\begin{figure*}[t]
\centering
\tiny
\setlength{\tabcolsep}{2.2pt}
\renewcommand{\arraystretch}{0.95}

\begin{minipage}[t]{0.45\linewidth}
\vspace{-0.2pt}
\centering
\scriptsize
\captionsetup{type=table}
\captionof{table}{\textbf{Baselines and adaptation paradigms.}
Detection reports COCO-mapped mAP, classification reports mapped VLM top-1 accuracy, and retrieval reports degraded-to-pristine R@1. FT = fine-tune, DA = domain adaptation, FR = feature recovery, PR = pixel restoration.}
\label{tab:detection-baselines}
\vspace{-0.45em}

{\fontsize{10pt}{10pt}\selectfont
\scalebox{0.73}[0.67]{
\rowcolors{2}{gray!7}{white}
\begin{tabular}{llc}
\toprule
\rowcolor{gray!22}
\textbf{Paradigm} & \textbf{Method} & \textbf{Score}$\uparrow$ \\
\midrule
\multicolumn{3}{l}{\textit{Degraded-object detection, mAP@0.5:0.95}} \\
No adapt.    & Base RF-DETR\cite{rf-detr}   & 0.338 \\
PR           & Restormer\cite{Zamir2021Restormer} + RF-DETR          & 0.375 \\
PR           & WildIR\cite{wildir} + RF-DETR   & 0.266 \\
DA           & SFA~\cite{wang2021exploring} & 0.161 \\
DA           & DATR~\cite{chen2025datr}     & 0.293 \\
FR           & Base + \module               & 0.442 \\
Full FT      & RF-DETR FT                   & 0.572 \\
\rowcolor{gray!12}
FT + FR    & RF-DETR FT + \module         & \textbf{0.647} \\
\midrule
\multicolumn{3}{l}{\textit{Pristine-State Recovery, top-1 accuracy (\%)}} \\
PR           & WildIR\cite{wildir} + Qwen3-VL     & 36.44 \\
PR           & WildIR + InternVL3.5  & 22.57 \\
\midrule
\multicolumn{3}{l}{\textit{Degraded-to-pristine retrieval, R@1}} \\
PR           & Restormer + Qwen3-VL        & 74.55 \\
FR           & Qwen3-VL + \module                      & \textbf{85.28} \\
PR           & Restormer + LLaVA-OV-1.5    & 80.97  \\
FR           & LLaVA-OV-1.5 + \module                  & \textbf{86.09} \\
\bottomrule
\end{tabular}
}
}
\end{minipage}
\hfill
\begin{minipage}[t]{0.53\linewidth}
\vspace{0pt}
\centering

\captionsetup{type=figure,width=\linewidth}
\includegraphics[
    width=0.85\linewidth,
    height=0.3\textheight,
    keepaspectratio
]{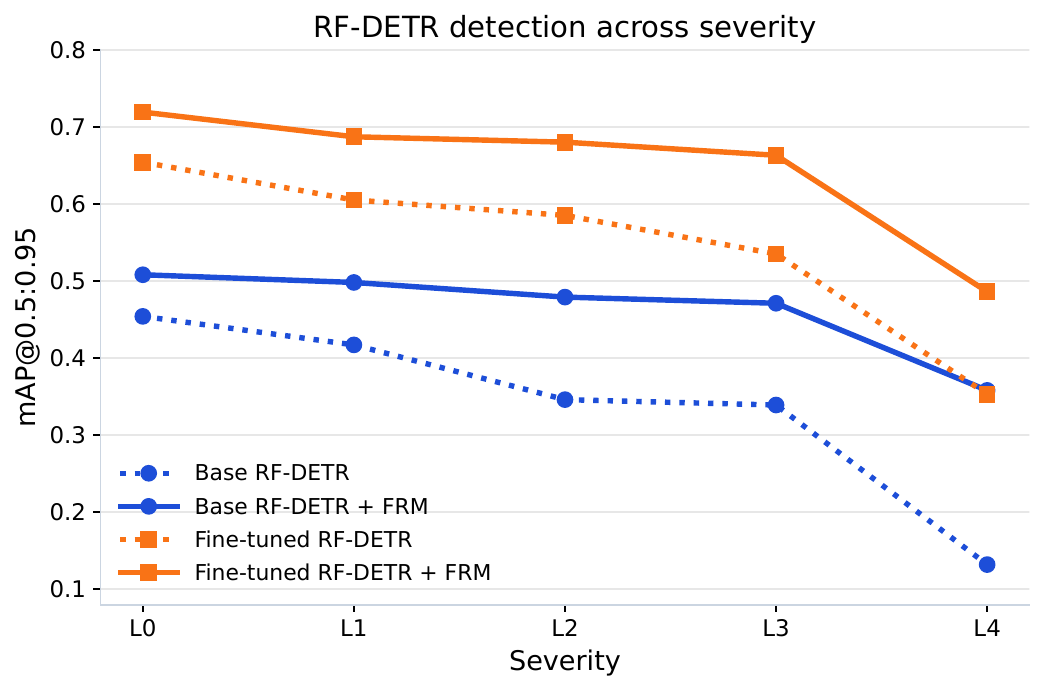}
\vspace{-0.65em}
\captionof{figure}{\textbf{RF-DETR detection across severity.}
\module improves both frozen and fine-tuned RF-DETR.}
\label{fig:rep-recovery}

\vspace{-0.35em}

\scriptsize
\captionsetup{type=table,width=\linewidth}
\captionof{table}{\textbf{Frozen encoder feature recovery.}
Degraded-to-pristine recovery on object-level test set (Base $\rightarrow$ + \module ). }
\label{tab:encoder-recovery}
\vspace{-0.25em}

\resizebox{0.95\linewidth}{!}{
\rowcolors{2}{gray!7}{white}
\begin{tabular}{lcccc}
\toprule
\rowcolor{gray!22}
\textbf{Encoder} & \textbf{MSE}$\downarrow$ & \textbf{Cos.}$\uparrow$ & \textbf{R@1}$\uparrow$ & \textbf{R@5}$\uparrow$ \\
\midrule
CLIP ViT-L/14 ~\cite{radford2021learning}
& 0.6719 $\rightarrow$ \textbf{0.4686}
& 0.6537 $\rightarrow$ \textbf{0.7492}
& - & - \\
SigLIP2~\cite{siglip2}
& 2.4172 $\rightarrow$ \textbf{1.3693}
& 0.6045 $\rightarrow$ \textbf{0.7421}
& 24.2 $\rightarrow$ \textbf{39.0}
& 77.3 $\rightarrow$ \textbf{96.2} \\
\bottomrule
\end{tabular}
}
\end{minipage}

\vspace{-1.5em}
\end{figure*}

\boldheader{Detection on \benchmark scenes. }Table~\ref{tab:detection-baselines} compares adaptation paradigms on the scene-level test set, with severity trends shown in Figure~\ref{fig:rep-recovery}. \module improves the frozen RF-DETR by 30.8\% and adds a further 13.1\% on top of the fine-tuned host while preserving pristine performance within 0.001 mAP. The additive gain indicates that fine-tuning and \module address different failure modes. Fine-tuning broadens the encoder's input distribution to include degraded images, while \module corrects residual feature misalignment that persists after fine-tuning. Statistics-only domain adaptation methods, including SFA and DATR, underperform the frozen baseline, confirming that marginal feature alignment is insufficient when degradation produces structured, per-instance drift requiring paired correction. Pixel restoration is insufficient for this setting. Restormer gives only a small detection gain (0.338$\rightarrow$0.375 mAP), while WildIR reduces RF-DETR performance to 0.266 mAP. WildIR also yields low pristine-state object classification accuracy with frozen VLMs (0.364 on Qwen3-VL and 0.225 on InternVL3.5), suggesting that visually restoring local image quality does not reliably recover the pre-degradation identity cues needed for recognition.



\boldheader{Feature-space recovery generalizes across encoder families.}
To isolate feature-space recovery from detector-specific confounds, we apply the same FRM recipe to frozen CLIP and SigLIP2 features (Table~\ref{tab:encoder-recovery}). Despite different pretraining objectives, contrastive for CLIP and sigmoid contrastive for SigLIP2, \module consistently moves degraded features toward paired pristine targets. It reduces MSE by 30.3\% on CLIP and 43.4\% on SigLIP2, with corresponding cosine similarity gains of +0.095 and +0.138. The effect carries through to pristine-state retrieval, where \module improves SigLIP2 degraded-to-pristine Top-1 by $+14.8$ points and Top-5 by $+18.9$ points, with Top-5 nearly reaching the pristine-query ceiling of $100\%$. These gains support the view that fire-induced degradation produces structured displacement on the feature manifold across encoder families.

\vspace{-0.8em}
\subsection{Pristine-State Recovery in VLMs}
\label{sec:exp-vlm}

\begin{figure}[t]
    \centering
    \includegraphics[width=\linewidth, height=0.2\linewidth]{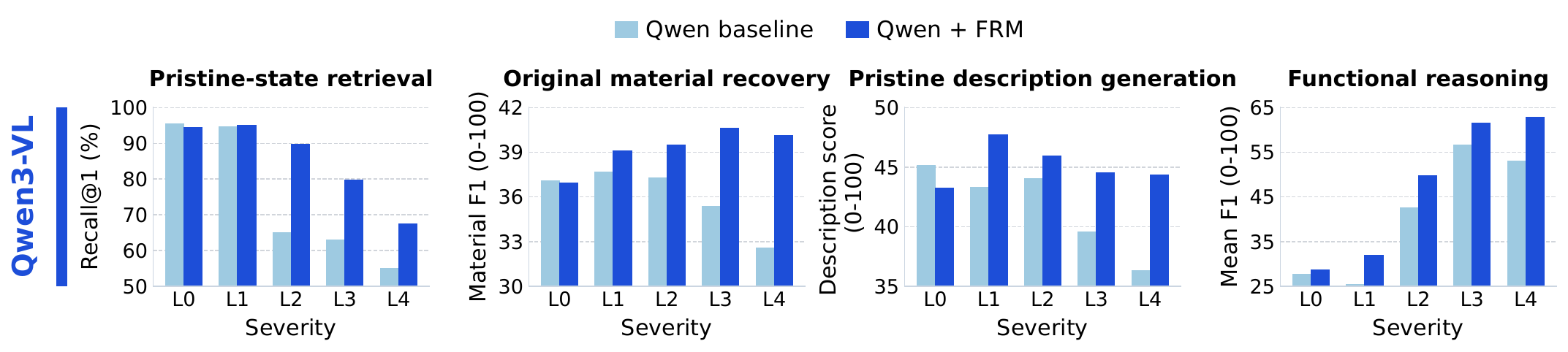}
    \caption{\textbf{Per-level Qwen3-VL pristine-state recovery.}
    Baseline is shown in light color and \module in dark color across L0-L4. Columns report pristine-state retrieval R@1, original material recovery F1, pristine-description score, and functional-reasoning mean F1. \module gains are strongest at higher severity.}
    \label{fig:vlm-severity}
    \vspace{-0.5em}
\end{figure}
\vspace{-0.5em}
\begin{table*}[t]
\centering
\tiny
\setlength{\tabcolsep}{2pt}
\renewcommand{\arraystretch}{0.72}

\begin{minipage}[t]{0.50\linewidth}
\vspace{0pt}
\centering
\tiny
\setlength{\tabcolsep}{2pt}
\renewcommand{\arraystretch}{0.72}
\captionsetup{type=table,width=\linewidth}
\captionof{table}{\textbf{VLM task performance across five frozen hosts.} Base is mean performance of frozen VLM over L0-L4; $\Delta$ is the gain from \module. Metrics are reported on a 0-100 scale, with per-level results in Appendix~\ref{app:vlm-per-level}.}
\label{tab:vlm-main}
\vspace{0.15em}

\rowcolors{2}{gray!8}{white}
\begin{tabularx}{\linewidth}{@{}>{\raggedright\arraybackslash}X
                                >{\centering\arraybackslash}p{0.18\linewidth}
                                >{\centering\arraybackslash}p{0.18\linewidth}
                                >{\centering\arraybackslash}p{0.18\linewidth}@{}}
\toprule
\rowcolor{gray!22}
\textbf{Model} & \textbf{Base} & \textbf{+\module} & \textbf{$\Delta$} \\
\midrule

\rowcolor{white}
\multicolumn{4}{@{}l}{\textit{\textbf{Task:} Pristine-state retrieval (R@1)}} \\
Qwen3-VL     & 74.55 & 85.28 & $+10.73$ \\
Qwen2.5-VL   & 60.20 & 72.03 & $+11.83$ \\
InternVL3.5  & 59.09 & 74.92 & $+15.83$ \\
Molmo2       & 85.06 & \textbf{90.00} & $+4.94$ \\
LLaVA-OV-1.5 & 83.97 & 86.09 & $+2.12$ \\
\midrule

\rowcolor{white}
\multicolumn{4}{@{}l}{\textit{\textbf{Task:} Original material recovery (micro-F1)}} \\
Qwen3-VL     & 36.00 & \textbf{39.24} & $+3.24$ \\
Qwen2.5-VL   & 30.50 & 36.78 & $+6.28$ \\
InternVL3.5  & 26.10 & 35.91 & $+9.81$ \\
Molmo2       & 23.56 & 27.52 & $+3.96$ \\
LLaVA-OV-1.5 & 22.31 & 26.89 & $+4.58$ \\
\midrule

\rowcolor{white}
\multicolumn{4}{@{}l}{\textit{\textbf{Task:} Pristine description generation (description score)}} \\
Qwen3-VL     & 41.67 & \textbf{45.15} & $+3.48$ \\
Qwen2.5-VL   & 40.50 & 44.00 & $+3.50$ \\
InternVL3.5  & 27.01 & 35.42 & $+8.41$ \\
Molmo2       & 33.80 & 38.13 & $+4.33$ \\
LLaVA-OV-1.5 & 35.48 & 39.34 & $+3.86$ \\
\midrule

\rowcolor{white}
\multicolumn{4}{@{}l}{\textit{\textbf{Task:} Functional reasoning (mean micro-F1)}} \\
Qwen3-VL     & 41.02 & 46.96 & $+5.94$ \\
Qwen2.5-VL   & 38.11 & 43.62 & $+5.51$ \\
InternVL3.5  & 36.13 & 40.48 & $+4.35$ \\
Molmo2       & 42.64 & \textbf{47.95} & $+5.31$ \\
LLaVA-OV-1.5 & 39.76 & 43.07 & $+3.31$ \\
\bottomrule
\end{tabularx}
\end{minipage}
\hfill
\begin{minipage}[t]{0.47\linewidth}
\vspace{0pt}
\centering
\tiny
\setlength{\tabcolsep}{2.2pt}
\renewcommand{\arraystretch}{0.82}
\captionsetup{type=table,width=\linewidth}
\captionof{table}{\textbf{Ablations.}
We ablate the recovery loss, cosine weight, and real-crop transfer. Metrics are cosine distance to pristine, identity accuracy, and retrieval R@1. Additional \module ablations are provided in Appendix~\ref{app:frm-ablations}.}

\label{tab:ablations}
\vspace{0.15em}

\rowcolors{3}{gray!8}{white}
\begin{tabularx}{\linewidth}{@{}>{\raggedright\arraybackslash}X
                                >{\centering\arraybackslash}p{0.19\linewidth}
                                >{\centering\arraybackslash}p{0.19\linewidth}
                                >{\centering\arraybackslash}p{0.19\linewidth}@{}}
\toprule
\rowcolor{gray!22}
\textbf{Configuration} & \textbf{Cos.$\downarrow$} & \textbf{ID$\uparrow$} & \textbf{R@1$\uparrow$} \\
\midrule
\rowcolor{white}
\multicolumn{4}{@{}l}{\textbf{Ablation 1: }\textit{Loss formulation, Qwen3-VL post-merger}} \\
No \module                    & 0.260 & 32.69 & 74.55 \\
LN-MSE                        & 0.030 & 36.47 & 57.83 \\
MSE only                      & 0.027 & 36.81 & 83.17 \\
Huber + $\mathcal{L}_{\cos}$  & 0.026 & 37.08 & 84.63 \\
MSE + 0.1$\mathcal{L}_{\cos}$ & \textbf{0.025} & \textbf{37.31} & \textbf{85.28} \\
\midrule

\rowcolor{white}
\multicolumn{4}{@{}l}{\textbf{Ablation 2: }\textit{Cosine weight $\lambda$, MSE+$\lambda\mathcal{L}_{\cos}$}} \\
No \module                    & 0.260 & 32.69 & 74.55 \\
$\lambda=0.01$ & 0.026 & 36.94 & 84.12 \\
\rowcolor{gray!12}
$\lambda=0.10$ & 0.025 & \textbf{37.31} & \textbf{85.28} \\
$\lambda=0.50$ & \textbf{0.024} & 37.14 & 84.37 \\
$\lambda=1.00$ & 0.027 & 36.54 & 82.76 \\
\midrule

\rowcolor{gray!22}
\textbf{Real crops} & \textbf{Base} & \textbf{+\module} & \textbf{$\Delta$} \\
\midrule
\rowcolor{white}
\multicolumn{4}{@{}l}{\textbf{Ablation 3: }\textit{Real-crop transfer, Qwen3-VL}} \\
\rowcolor{white}
\multicolumn{4}{@{}l}{\textit{L2-L4 average:}} \\
Retrieval R@1                 & 49.16 & 53.03 & +3.87 \\
Material F1                   & 25.31 & 29.98 & +4.67 \\
Description gen.              & 27.47 & 32.53 & +5.06 \\
Functional reasoning          & 29.31 & 33.80 & +4.49 \\
\midrule

\rowcolor{white}
\multicolumn{4}{@{}l}{\textit{L4 only}:} \\
Retrieval R@1                 & 43.28 & 46.03 & +2.75 \\
Material F1                   & 22.47 & 26.88 & +4.41 \\
Description gen.              & 24.93 & 27.84 & +2.91 \\
Functional reasoning          & 26.84 & 30.63 & +3.79 \\
\bottomrule


\end{tabularx}
\end{minipage}
\vspace{-0.5em}
\end{table*}
\vspace{-0.4em}

\boldheader{Retrieval gains concentrate at higher severity. }\module improves degraded-to-pristine Recall@1 across all evaluated VLM hosts (Table~\ref{tab:vlm-main}, Figure~\ref{fig:vlm-severity}; full per-level results in Table~\ref{tab:vlm-per-level-full}). Averaged across hosts, retrieval gains are nearly flat at mild degradation ($+0.73$ points over L0-L1) but rise sharply at higher severity ($+14.66$ points over L2-L4). This pattern argues against a fixed correction bias and is consistent with \module correcting feature displacement that grows with physical degradation. The gains are larger for weaker retrieval baselines and smaller for stronger ones, suggesting that \module recovers pristine-state signal that weaker encoders lose while stronger encoders partially retain it without correction.

\boldheader{Lower baseline performance leaves more room for recovery.} The largest absolute gains on original material recovery and pristine description generation occur on InternVL3.5 (Table~\ref{tab:vlm-main}), which has the weakest degraded-input baseline on these tasks (material $+9.81$ points, description generation $+8.41$ points). \module's benefit is better explained by degradation sensitivity than by model capacity. It helps most when the frozen host loses more pristine-state information under physical transformation. Operationally, this suggests that \module is most valuable for host-task pairs with large degradation-induced performance drops.

\boldheader{Pristine alignment transfers to functional reasoning. }\module improves the functional reasoning combined score on every host (Table~\ref{tab:vlm-main}), with a mean aggregate gain of $+4.9$ points across VLMs and larger gains at severe degradation levels (average $+7.6$ at L3-L4, Figure~\ref{fig:vlm-severity}). \module is trained only on paired features with no per-image task labels, yet recovered features support reasoning about pre-degradation function and restoration needs. This suggests that the frozen language models already encode object-function knowledge, and \module restores the visual evidence needed to invoke it. Without \module, models often fall back to coarse functional guesses such as ``seating,'' or ``container,'' which can partially match broad tags without recovering the specific object. The high-severity gains suggest \module recovers object-specific function rather than improving generic guessing.

\vspace{-0.5em}
\subsection{Ablations}
\label{sec:exp-ablations}
\vspace{-0.5em}
We ablate the recovery loss and test real-crop transfer under moderate and severe degradation, where \module is most useful (Table~\ref{tab:ablations}, with additional ablations in Appendix~\ref{app:frm-ablations} and Table~\ref{tab:app-frm-ablations}). \textbf{i. Loss formulation.} All loss ablations use the Qwen3-VL post-merger tap. LN-MSE reduces cosine distance but drops retrieval below the no-\module baseline, showing that cosine distance alone is not sufficient and that token magnitude matters for pooled retrieval. MSE alone restores retrieval, while MSE$+\mathcal{L}_{\cos}$ gives the best tradeoff across cosine distance, identity accuracy, and R@1. Huber$+\mathcal{L}_{\cos}$ is close but does not exceed it, suggesting that the quadratic penalty on large deviations helps severe degradation tokens. \textbf{ii. Cosine weight.} R@1 peaks at $\lambda=0.1$ and changes smoothly around it. Although $\lambda=0.5$ gives the lowest cosine distance, $\lambda=0.1$ gives the best identity accuracy and retrieval. Across $\lambda \in \{0.01,0.1,0.5\}$, R@1 varies by only 1.16 points, so the joint loss is robust to the exact weight in this range. We use MSE$+0.1\mathcal{L}_{\cos}$ throughout. \textbf{iii. Real-crop transfer.} We evaluate Qwen3-VL on 500 real post-fire  crops from classes present in \benchmark at L2-L4, since real post-fire imagery is concentrated in moderate to severe damage and contains few reliable L0-L1 analogs. \module improves every real-crop task on average and remains positive at L4, where near-total destruction leaves less recoverable evidence.

\vspace{-0.7em}
\section{Conclusion \& Discussion}
\label{sec:discussion}
\vspace{-0.5em}
\boldheader{Conclusion. }We introduced \benchmark, a transformation-aware benchmark for post-fire object understanding, comprising 21.4K real-image-grounded synthetic scenes and object-level pristine-to-degraded trajectories spanning 499 object identities across 189 categories. We also introduced \modulefull (\module), a plug-and-play residual module that maps degraded features toward their pristine counterparts in a frozen feature space, enabling downstream models to reason about damaged objects in terms of their pre-degradation state. Across eight frozen hosts spanning RF-DETR, CLIP, SigLIP2, and five VLMs, \module improves every task, with gains that concentrate at higher degradation severity, yielding 9.0–30.8\% relative improvement across detection, retrieval, and reasoning. Unlike prior robustness work that targets image-quality degradation or distributional shift, \benchmark and \module address irreversible physical transformations where the object itself has changed material state, lost parts, or retained only partial structural evidence. This positions the framework as a step toward real-world post-fire applications, from identifying hazardous objects during emergency response to inventorying destroyed property for insurance claims and reconstructing pre-incident scene contents for investigation.

\boldheader{Limitations and Future Directions.}
The recovery direction is task-dependent: \module recovers pristine-aligned features, which improves queries about original identity, materials, function, and restoration, but removes exactly the damage signal needed for queries about current state (e.g., ``describe the fire damage''). \module should be disabled for such queries, and a query-aware gating mechanism would allow a single host to serve both pristine-state and damage-state queries. The module is also tied to a specific (encoder, tap, feature space) configuration, so each new host requires a separately trained module. Adding a new object identity requires paired supervision from a pristine visual anchor and at least three degraded observations across severity levels; relaxing this requirement is an open direction. Real-crop gains remain positive but smaller than synthetic gains, reflecting residual synthetic-to-real gaps from generative-model limits, approximate real-image severity labels, cluttered post-fire conditions, and material failure modes not fully captured by generation. Finally, identity recovery is only one of the cues humans use under severe degradation: room layout, neighboring objects, and surviving structural parts all carry information that \module currently ignores. Incorporating spatial context and part-level understanding is a natural next step, as is extending the paired-supervision recipe to other irreversible transformations such as flood damage, structural collapse, and weathering, wherever pre-event visual references exist.

\textbf{Impact Statement. }This work targets visual recognition under irreversible physical degradation, with applications in emergency response, forensic investigation, and insurance assessment. Reliable detection and identification of objects in post-fire environments could support firefighters in locating hazardous items in collapsed structures, assist investigators in reconstructing pre-incident scene contents, and reduce the manual effort required to inventory damaged property after disasters. Releasing \benchmark and the \module training pipeline lowers the barrier for further research on degraded-object understanding, including extensions to other irreversible transformations such as flood damage and weathering. The deployment risks are largely shared with general-purpose object recognition. Pristine-state predictions are inferences from incomplete physical evidence and can be wrong, particularly at high severity; using them as evidence in legal or insurance contexts requires uncertainty quantification and human oversight that this work does not provide. The benchmark consists of synthetic scenes grounded in real post-fire imagery; while we filter for realism, distributional gaps to specific deployment environments remain, and downstream users should validate on in-domain data before operational use. We are not aware of dual-use concerns specific to this work.

\bibliographystyle{plainnat}
\bibliography{neurips_2026}

\clearpage
\appendix
\section*{Appendix}
\label{app:roadmap}

The appendix is organized as follows:
\begin{itemize}
    \item \textbf{Appendix~\ref{app:dataset-details}} reports dataset statistics for the object-level and scene-level.
    \item \textbf{Appendix~\ref{app:dataset}} describes the object and scene generation pipeline, including prompts, real-image grounding, and filtering.
    \item \textbf{Appendix~\ref{app:task-details}} provides the evaluation prompts and scoring protocols for all tasks.
    \item \textbf{Appendix~\ref{app:frm-hosts}} details \module instantiations across VLM and frozen-encoder hosts.
    \item \textbf{Appendix~\ref{app:vlm-per-level}} reports full per-level VLM results and frozen-encoder recovery results.
    \item \textbf{Appendix~\ref{app:frm-ablations}} provides additional ablations on capacity, integration, and tap-point selection.
\end{itemize}


\section{Dataset Statistics}
\label{app:dataset-details}

This appendix summarizes the subset statistics, object and scene coverage, and degradation diagnostics for \benchmark. Detailed generation prompts, grounding-judge criteria, and filtering protocol are provided in Appendix~\ref{app:dataset}.

\subsection{Subset Statistics}
\label{app:subset-stats}

\begin{figure}[ht]
\centering
\includegraphics[width=\linewidth]{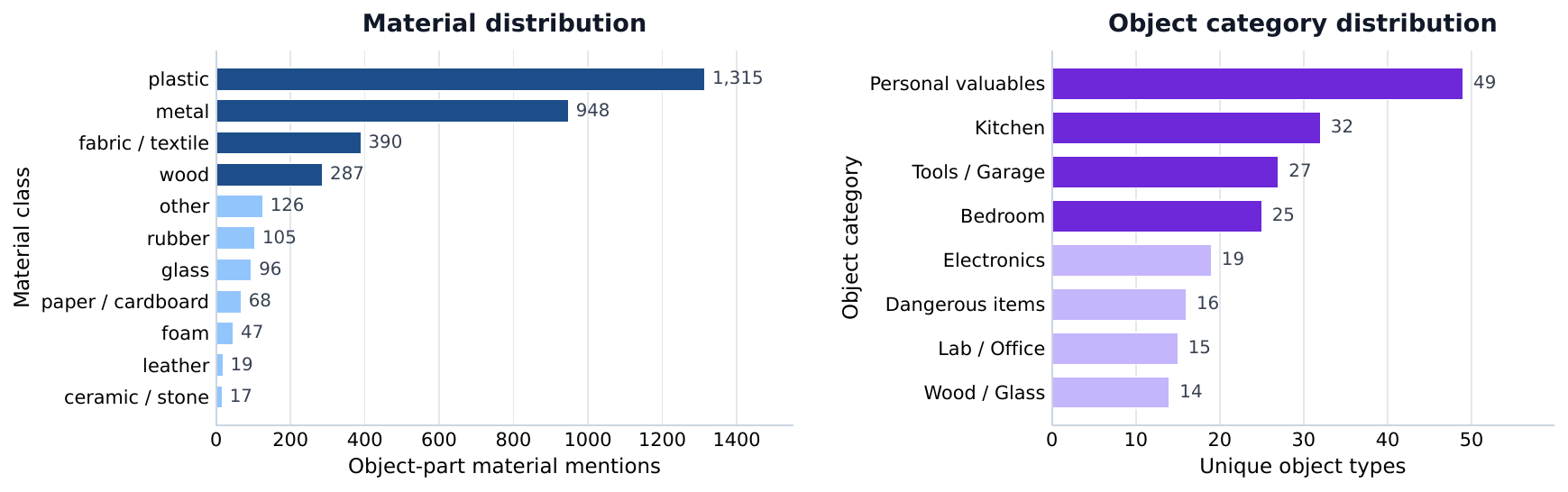}
\caption{\textbf{Object inventory statistics.}
The object inventory spans 499 object identities, 189 object types, and 11 normalized material classes. Materials are long-tailed, with plastic, metal, fabric/textile, and wood most represented.}
\label{fig:appendix-object-inventory}
\end{figure}

\begin{figure}[ht]
\centering
\includegraphics[width=\linewidth, height=0.37\linewidth]{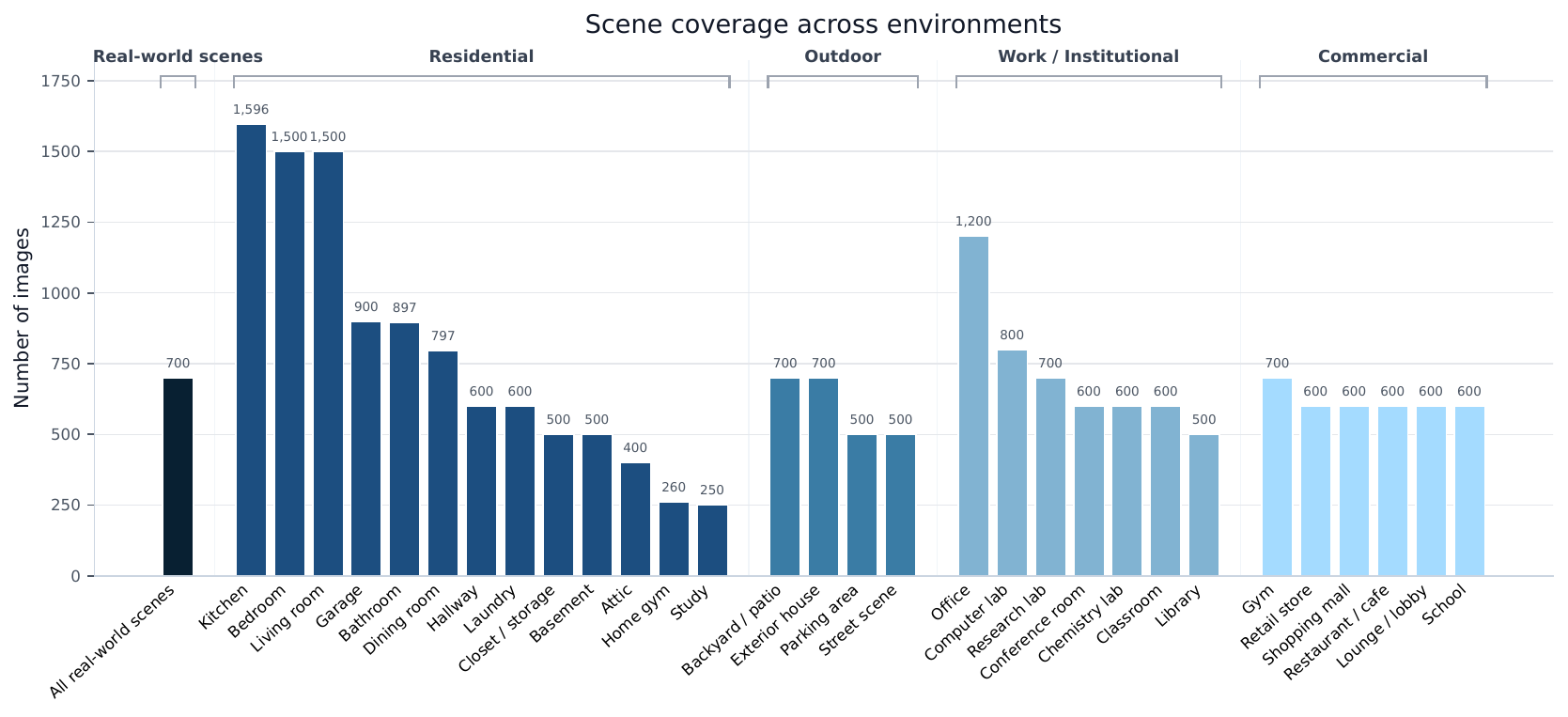}
\caption{\textbf{\benchmark scene coverage.}
\benchmark contains 21{,}400 accepted synthetic post-fire scenes grounded by approximately 700 real post-fire reference images. Scenes span residential, outdoor, work/institutional, and commercial environments.}
\label{fig:dataset_composition}
\end{figure}

\begin{figure}[ht]
\centering
\includegraphics[width=\linewidth]{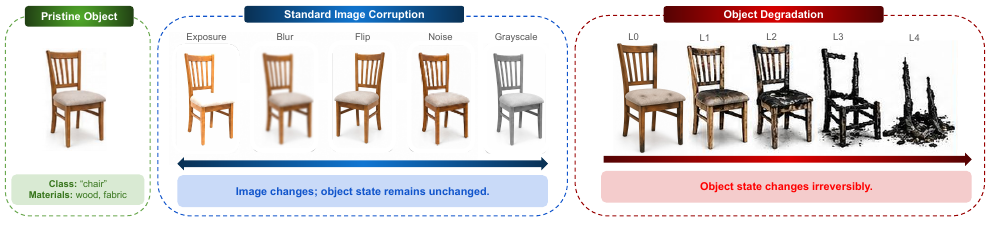}
\caption{\textbf{Physical transformation is not image corruption.}
Standard corruptions alter image appearance while preserving object state. Post-fire degradation changes material and structure, requiring recovery of pre-degradation identity and properties.}
\label{fig:corruption_vs_degradation}
\end{figure}

\begin{figure}[ht]
    \centering
    \includegraphics[width=\linewidth]{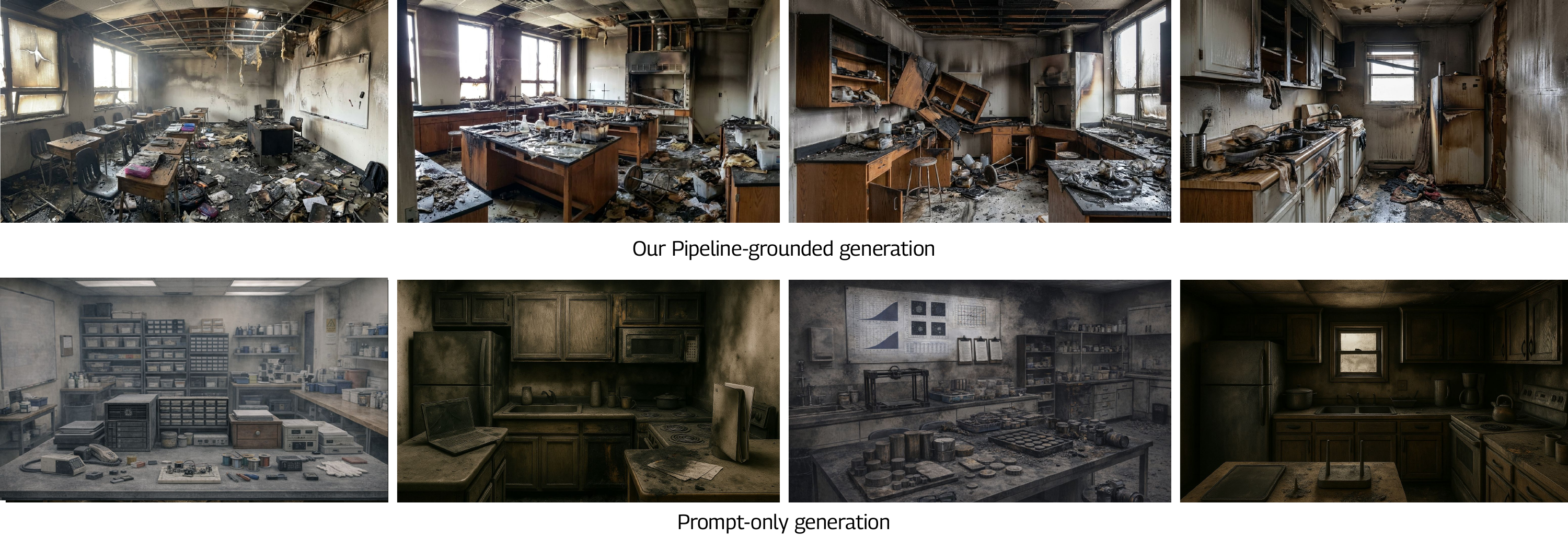}
    \caption{\textbf{Effectiveness of the scene-generation pipeline.}
    Both rows use Gemini~3 Pro Image as the generator. The top row shows scenes produced by our pipeline, which conditions generation on 2-4 real post-fire reference images and applies a VLM judge for accept/reject filtering. The bottom row uses the same scene prompts but removes real-image grounding and judge-based filtering. Reference grounding and filtering improve scene realism, object grounding, and post-fire plausibility.}
    \label{fig:generation-pipeline}
\end{figure}

Figures~\ref{fig:appendix-object-inventory} and~\ref{fig:dataset_composition} summarize object-level and scene-level coverage. Figure~\ref{fig:corruption_vs_degradation} contrasts standard image corruptions with physical transformation. Figure~\ref{fig:generation-pipeline} shows the effect of real-image grounding and VLM filtering.

\subsection{Degradation Severity Metadata}
\label{app:severity-metadata}

Each object trajectory is annotated with approximate severity metadata for soot coverage, surface or material failure, deformation, and structural integrity. These ranges guide generation and support severity-stratified evaluation. Each object trajectory is annotated with approximate severity metadata for soot coverage, surface or material failure, deformation, and structural integrity (Table~\ref{tab:degradation-metrics}).

\begin{table}[b]
\centering
\scriptsize
\setlength{\tabcolsep}{3pt}
\caption{\textbf{Degradation severity metrics by level.}
Approximate ranges used as object-level generation metadata across the five degradation levels.}
\label{tab:degradation-metrics}
\begin{tabular}{lccccc}
\toprule
Metric & L0 & L1 & L2 & L3 & L4 \\
\midrule
Soot coverage (\%) & 30-45 & 45-60 & 60-75 & 72-87 & 87-96 \\
Surface/material failure (\%) & 30-50 & 30-45 & 25-40 & 40-60 & 60-80 \\
Deformation (\%) & 5-15 & 15-30 & 35-55 & 55-75 & 75-92 \\
Structural integrity (\%) & 85-95 & 65-82 & 40-58 & 15-35 & 4-13 \\
\bottomrule
\end{tabular}
\end{table}

\subsection{Annotation Summary}
\label{app:annotation-summary}

Object-level metadata includes part-level material composition, object-specific failure notes, and per-level degradation descriptions. Scene-level metadata includes bounding boxes, mapped object labels, visible materials, degradation states, scene captions, and generation metadata. Intended object and material inventories from the generation prompt are stored as metadata and are not used as ground truth for evaluation.

\section{Generation Prompts and Filtering}
\label{app:dataset}

This appendix provides the prompt templates used to generate the scene-level and object-level subsets of \benchmark. All generation uses Gemini~3 Pro Image for image synthesis and Gemini~3 Flash as the grounding judge.

\begin{tcolorbox}[
title={Object Trajectory Metadata Example: Dresser},
breakable,
colback=white,
colframe=black!60
]

\begin{lstlisting}[style=jsonTiny]
{
  "object": "dresser",
  "scene_category": "bedroom",
  "trajectory": "1 pristine + 5 degraded states",
  "split_unit": "object identity",
  "degradation_levels": ["L0", "L1", "L2", "L3", "L4"]
}
\end{lstlisting}

\vspace{2pt}
{\centering
\includegraphics[width=0.72\linewidth]{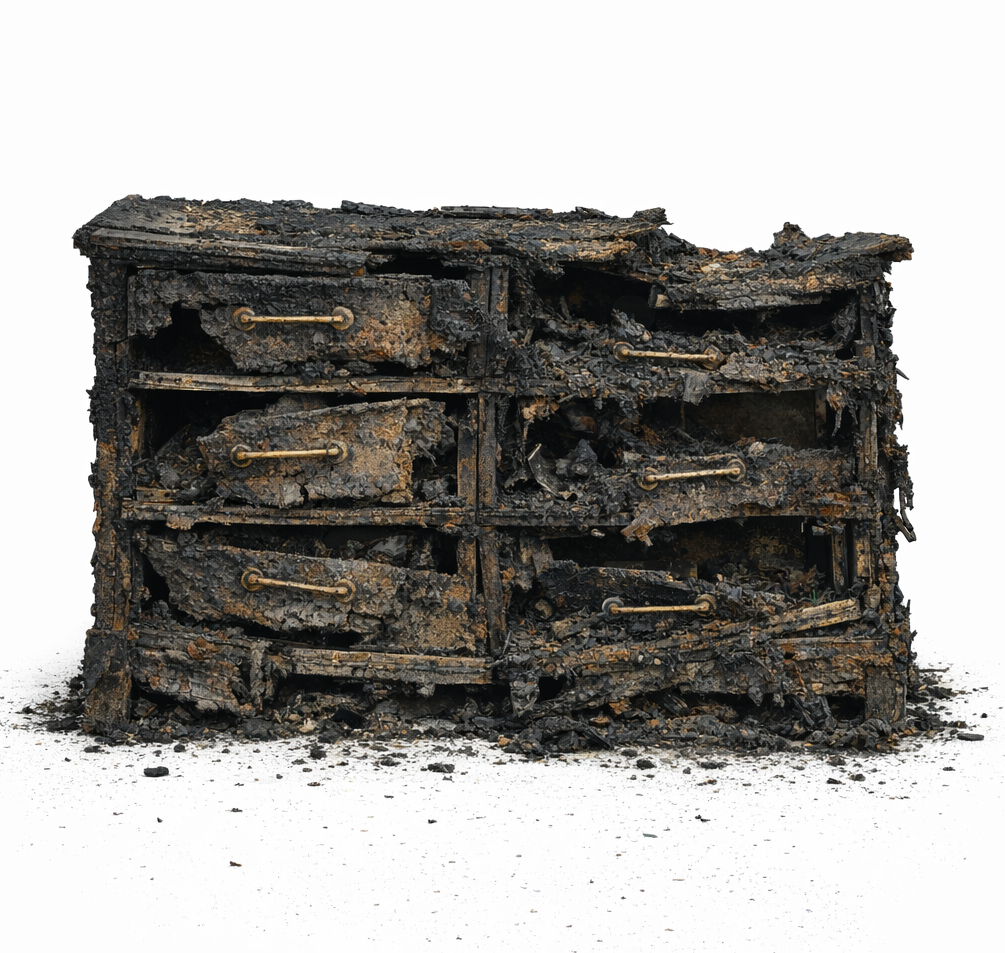}
\par}
\vspace{2pt}

\begin{lstlisting}[style=jsonTiny]
{
  "part_materials": [
    {"part": "top surface", "material": "MDF or hardwood with veneer"},
    {"part": "body panels", "material": "MDF laminate or hardwood"},
    {"part": "drawer fronts", "material": "MDF or hardwood"},
    {"part": "drawer boxes", "material": "MDF or plywood"},
    {"part": "drawer runners", "material": "steel slides"},
    {"part": "drawer pulls", "material": "brass, zinc alloy, or ABS plastic"},
    {"part": "base plinth", "material": "hardwood or MDF"},
    {"part": "back panel", "material": "thin MDF or hardboard"},
    {"part": "finish coating", "material": "polyurethane lacquer or paint"}
  ]
}
\end{lstlisting}

\vspace{2pt}

\begin{lstlisting}[style=jsonTiny]
{
  "material_failure_notes": {
    "wood_MDF": "finish blisters and peels; MDF chars, swells, and crumbles",
    "hardware": "steel drawer slides and metal pulls survive longest",
    "late_stage": "dresser collapses into a charred structure with surviving hardware"
  }
}
\end{lstlisting}

\vspace{2pt}

\begin{lstlisting}[style=jsonTiny]
{
  "severity_ladder": {
    "L0": "heavy soot, finish blistering, early surface char, structure intact",
    "L1": "deep surface char, cracking, stronger heat damage",
    "L2": "burn-through in weaker members, exposed hardware, partial failure",
    "L3": "collapsing charred skeleton with surviving metal components",
    "L4": "near-total destruction with weak identity cues in hardware and debris"
  }
}
\end{lstlisting}

\end{tcolorbox}

\subsection{Object-Level Generation}
\label{app:object-prompts}

Each object trajectory is generated iteratively. The pristine image is generated first from a text-only prompt, and each subsequent degradation stage (L0-L4) is conditioned on the previous stage's image. This ensures cumulative, physically grounded damage progression.

\paragraph{Pristine prompt.} The pristine stage uses a studio product-photo prompt with explicit part-level material specifications and variant identity constraints to ensure distinct physical instances across variants of the same category.

\begin{tcolorbox}[colback=gray!5, colframe=gray!85, breakable, title={\small\textbf{Pristine Prompt Template}}]
\small
Single-object studio product photograph of a \texttt{\{obj\}}.

This exact object instance must be: \texttt{\{variant\_style\}}.\\
Design language: \texttt{\{variant\_manufacturing\}}.\\
Color / finish direction: \texttt{\{variant\_colorway\}}.

The object is centered, fills roughly 70 percent of the frame, fully visible, isolated on a pure white seamless background. Soft even studio lighting, sharp focus, realistic shadows only under the object, no stylization, no watermark, no text overlay.

Exact part-level construction that must be visible:\\
\texttt{\{part\_materials\}}

Hard constraints:
\begin{itemize}[leftmargin=*, itemsep=1pt, topsep=2pt]
\item Keep the object in a canonical product-photo orientation.
\item Keep background unchanged across all stages.
\item Do not add hands, props, shelves, tables, walls, flames, smoke clouds, or extra objects.
\item Show realistic material texture and manufacturing details.
\item This image must depict a DIFFERENT real-world variant than other versions of the same object category, not merely a recolor.
\item The silhouette, proportions, detailing, and subtype cues should clearly distinguish this instance from sibling variants.
\end{itemize}
\end{tcolorbox}

\paragraph{Degradation stage prompt.} Each degradation stage is conditioned on the previous stage's image and includes material-specific failure behavior, stage-level severity metrics, and strict constraints ensuring cumulative and spatially uneven damage.

\begin{tcolorbox}[colback=gray!5, colframe=gray!85, breakable, title={\small\textbf{Degradation Stage Prompt Template}}]
\small
Generate the next degradation image for a \texttt{\{obj\}} at stage \texttt{\{stage\_level\}} using the provided reference image(s) as the SAME physical object instance.

\texttt{\{variant\_identity\_block\}}

\texttt{\{consistency\_constraints\}}

\texttt{\{global\_realism\_rules\}}

Exact part-level construction that must remain physically grounded:\\
\texttt{\{part\_materials\}}

Material-specific failure behavior that MUST be obeyed per part:\\
\texttt{\{material\_behavior\}}

Object-specific failure notes:\\
\texttt{\{object\_specific\_failure\_notes\}}

Stage description - THIS IS THE TARGET STATE:\\
\texttt{\{stage\_description\}}

Critical image constraints:
\begin{itemize}[leftmargin=*, itemsep=1pt, topsep=2pt]
\item Make stage \texttt{\{stage\_level\}} CLEARLY and DRAMATICALLY different from the previous stage.
\item Every visible surface must show fire exposure - no pristine areas anywhere.
\item Keep the white studio background unchanged.
\item Do not add flames, smoke plumes, sparks, people, tables, or debris outside the object footprint.
\item Damage must be spatially uneven and material-aware - not a uniform dark overlay.
\end{itemize}
\end{tcolorbox}

\paragraph{Consistency and realism constraints.} The following constraints are injected into every degradation stage prompt to enforce material-aware failure behavior.

\begin{tcolorbox}[colback=gray!5, colframe=gray!85, breakable, title={\small\textbf{Material Failure and Realism Rules}}]
\small
\textbf{Material-specific failure rates:}
\textit{Paper and cardboard.} Among the first materials to fail. Curl and brown at edges, then char progressively inward. Thin sheets fragment into ash flakes. Cardboard delaminates at glue seams before charring. Final state is gray-white ash with occasional partially-surviving fragments in shielded regions.

\textit{Fabric and textile.} Synthetic fabrics (polyester, nylon) melt, shrink, and fuse into hard irregular beads or films. Natural fabrics (cotton, linen) char and disintegrate into fragile ash. Blended fabrics show both behaviors simultaneously. Seams and hems often survive longer than flat panels. Collapse follows loss of structural fiber integrity.

\textit{Foam and padding.} Melts, shrinks, and collapses rapidly. Exposed foam recedes from edges inward, leaving skeletal frames or springs visible underneath. Polyurethane foam produces sticky, re-solidified residue. Memory foam chars into brittle black mass.

\textit{Leather.} Shrinks dramatically (up to 50\% area loss), curls at edges, and develops deep surface cracking. Grain pattern distorts but remains partially visible. Blackens progressively from tan to dark brown to black. Does not melt or produce liquid residue.

\textit{Wood and MDF.} Surface scorches first, producing alligator-skin char patterns. Deeper charring follows with visible char depth increasing per stage. MDF delaminates at adhesive layers. Structural wood cracks along grain lines. Final stages show brittle, crumbling charcoal with ash accumulation at the base. 

\textit{Plastics (ABS, polycarbonate, polypropylene, PVC).} Soften and deform at relatively low temperatures. ABS blisters and warps. Polycarbonate yellows, then browns, then deforms. Polypropylene melts and pools into irregular puddles. PVC blackens and becomes brittle. Re-solidified plastic forms irregular lumpy masses with frozen drip patterns. Thin-walled plastic containers collapse inward. 

\textit{Rubber and silicone.} Blacken and shrink. Rubber hardens and cracks into segments. Silicone maintains shape longer but eventually chars and crumbles. Tire-like rubber develops deep fissures before fragmenting. Rubber coatings on handles and grips peel and curl away from underlying metal or plastic. 

\textit{Metal (steel, aluminum, copper, iron).} Most durable common material. Steel heat-tints through straw, blue, and purple before developing surface oxide scale. Aluminum softens, sags, and can partially melt at extreme temperatures, losing structural rigidity. Copper and brass develop heavy green-black oxidation. Iron rusts rapidly in post-fire moisture. Thin sheet metal warps and buckles. Fasteners (screws, hinges) often survive intact and become identifying features. Metal does not become ash. 

\textit{Glass.} Shatters from thermal shock (rapid heating or cooling with water). Tempered glass produces small granular fragments. Plate glass cracks in radiating patterns before falling out of frames. At extreme temperatures, glass may soften and slump but does not melt into liquid. Soot deposits on surviving glass surfaces. 

\textit{Ceramic and stone.} Ceramic glazes craze (develop fine crack networks) and may spall. Unglazed ceramic and stone crack under thermal stress but maintain overall form. Neither material melts. Porcelain can shatter from thermal shock. Stone surfaces accumulate soot and may show heat-induced color changes (reddening of limestone, darkening of granite). 

\textit{Electronics and composite materials.} PCBs char and delaminate, exposing copper traces. Solder melts and beads on surfaces. LCD and LED screens crack and blacken. Battery cells swell, rupture, and may leave chemical residue. Wire insulation melts away, exposing bare copper conductors. Integrated circuits and ceramic chip packages may survive as identifiable components. 

\textit{Gemstones.} Diamond, sapphire, ruby, and similar crystalline stones survive fire temperatures and remain as intact loose stones among debris. Settings (typically metal) may deform or melt around them. Softer stones (opal, turquoise) may crack or discolor. 

\textbf{Failure rate ordering (earliest to latest):} Paper and cardboard $\rightarrow$ thin fabric and foam $\rightarrow$ thick fabric and leather $\rightarrow$ wood and MDF $\rightarrow$ plastic and rubber $\rightarrow$ glass $\rightarrow$ ceramic and stone $\rightarrow$ metal $\rightarrow$ gemstones. 

\textbf{Stage consistency:} Damage must accumulate from the previous stage only. Each new stage must be clearly and dramatically worse. Viewpoint, framing, scale, and orientation are preserved until physical collapse makes it impossible. Do not invent parts not present in the pristine image. Different materials on the same object must fail at different rates, so a single object at intermediate stages will show a mix of material states (e.g., fabric seat cover charred while the metal frame beneath shows only heat-tinting).
\end{tcolorbox}

\paragraph{Degradation stage definitions.} 
Each object trajectory progresses through five degradation stages (L0-L4). The following definitions specify the target physical state at each stage and are injected into the stage prompt as \texttt{\{stage\_description\}}.

\begin{tcolorbox}[colback=gray!5, colframe=gray!85, breakable, title={\small\textbf{Degradation Stage Definitions}}]
\small
\textbf{L0 - Mild surface damage.} Early-stage fire exposure. Light soot deposits on exposed surfaces. Minor discoloration or browning of heat-sensitive materials (paper, thin fabric). Plastic surfaces may show slight warping or surface bubbling. Metal shows no visible change or faint heat tinting. Object silhouette, proportions, and all parts remain fully intact. The object is immediately recognizable.

\textbf{L1 - Moderate surface damage.} Sustained heat exposure across all surfaces. Paper and thin fabric are partially charred or missing. Plastic shows visible warping, bubbling, or early melting. Wood surfaces show surface scorching and early char patterns. Metal begins to show heat-tint discoloration. Leather shrinks slightly and darkens. Foam begins to recede from edges. Object silhouette remains intact but surface appearance has changed substantially. Recognition requires attention to shape and remaining material cues.

\textbf{L2 - Structural onset.} Fire has begun to compromise structural integrity of weaker materials. Fabric and foam are largely gone or reduced to residue, exposing internal frames and components. Plastic has melted and re-solidified into deformed masses. Wood shows deep charring with visible char depth and cracking along grain. Rubber has hardened and cracked. Metal shows moderate oxidation and early warping of thin sections. Some parts may have detached or collapsed. Object is still recognizable from overall shape but individual material surfaces are heavily altered.

\textbf{L3 - Severe degradation.} Major structural failure of non-metallic materials. Wood is deeply charred and crumbling. Plastic has pooled or fused to other surfaces. Fabric, paper, foam, and leather are largely absent. Glass is shattered or missing. Metal components show heavy oxidation, soot coverage, and warping. Significant parts of the object may have collapsed, detached, or disintegrated. Object identity is recoverable primarily from surviving metal skeleton, overall proportions, and spatial arrangement of remaining components.

\textbf{L4 - Near-total destruction.} Approaching complete loss of original form. Non-metallic materials are reduced to ash, charred rubble, or re-solidified residue. Metal components are heavily deformed, oxidized, and may have partially melted (aluminum) or collapsed. Object silhouette is barely recognizable or unrecognizable without context. Identification requires expert-level inference from surviving fragments, hardware, and spatial cues. Some objects at this stage consist primarily of a debris pile with identifiable metal remnants.

\end{tcolorbox}

\subsection{Scene-Level Generation}
\label{app:scene-prompts}

Each scene is generated from a structured prompt conditioned on two to five real post-fire reference photographs. The prompt specifies scene type, damage level, material profile, clutter level, lighting, and socioeconomic band, with explicit instructions that references serve only as inspiration for realism and material fidelity.

\begin{tcolorbox}[colback=gray!5, colframe=gray!85, breakable, title={\small\textbf{Scene Generation Prompt Template (abbreviated)}}]
\small
You are generating a completely NEW, original real-world post-fire scene.

The reference images are ONLY for inspiration about realism, material degradation fidelity, structural damage realism, and scene complexity. Do NOT copy, recreate, or closely match any specific layout, room geometry, viewpoint, object arrangement, composition, or damage pattern from the references. The generated scene must differ significantly in geometry, camera angle, and object placement from all references.

Generate a realistic \texttt{\{scene\_type\}} after a fire incident. The image should be novel, physically grounded, hyper-realistic, and clearly distinct from the references.

\textbf{TARGET SCENE CONTROL:}
\begin{itemize}[leftmargin=*, itemsep=1pt, topsep=2pt]
\item Scene type: \texttt{\{scene\_type\}}
\item Damage level: \texttt{\{damage\_level\}}
\item Material profile: \texttt{\{material\_profile\}}
\item Clutter level: \texttt{\{clutter\_level\}}
\item Lighting mode: \texttt{\{lighting\}}
\item Socioeconomic / aesthetic band: \texttt{\{wealth\_style\}}
\item Layout style: \texttt{\{layout\_variant\}}
\end{itemize}

\textbf{PHYSICAL AND MATERIAL ACCURACY (CRITICAL):}
All visible materials must degrade according to real-world fire behavior.

\begin{itemize}[leftmargin=*, itemsep=1pt, topsep=2pt]
\item \textit{Wood:} charred, cracked, structurally weakened, sometimes partially collapsed. Load-bearing beams may sag or fracture. Exposed grain shows deep alligator-pattern charring. Ash accumulates at the base of vertical wooden surfaces. 
\item \textit{Plastic:} melted, warped, deformed, softened, pooled, or fused to adjacent surfaces. Thin-walled containers collapse inward. Re-solidified plastic forms irregular lumpy masses with frozen drip patterns on floors and shelving below. 
\item \textit{Fabric and upholstery:} burned away unevenly, collapsed, torn, or reduced to residue. Synthetic fabrics melt and fuse into hard beads or films. Natural fabrics char and fragment. Springs and frames become exposed as covering material fails. 
\item \textit{Foam and padding:} melted, collapsed, shrunken, missing in sections, or exposed beneath charred fabric. Polyurethane foam produces sticky re-solidified residue. Mattresses and cushions show deep cavities where foam has receded. 
\item \textit{Metal:} structurally present but heat-discolored, oxidized, soot-covered, or slightly warped. Steel shows straw-to-blue heat tinting. Aluminum may sag or partially melt. Thin sheet metal (appliance panels, ductwork) buckles and warps. Fasteners, hinges, and handles often survive intact. 
\item \textit{Glass:} shattered, fractured, partially missing, or heat-cracked. Window frames may retain jagged fragments. Tempered glass produces granular piles on the floor below. Mirrors crack in radiating patterns. Soot deposits on surviving glass surfaces. 
\item \textit{Paper and cardboard:} curled, charred, blackened, fragmented, reduced to ash, or partially surviving in shielded regions such as inside drawers or behind other objects. 
\item \textit{Rubber:} warped, softened, heat-damaged, partly melted, or degraded at edges. Cable insulation melts away exposing wire conductors. Rubber seals and gaskets shrink and crack. 
\item \textit{Concrete, tile, and masonry:} scorching, cracking, soot staining, and spalling where surfaces faced direct heat. Grout may crack and crumble. Floor tiles may lift or shatter from thermal expansion. Concrete walls show smoke staining gradients that indicate fire direction and intensity. 
\item \textit{Insulation and ceiling materials:} exposed fiberglass or mineral wool sags and discolors. Ceiling tiles collapse or hang partially detached. Drywall paper facing chars while the gypsum core remains as crumbling white residue.
\end{itemize}

\textbf{SPATIAL DAMAGE CONSISTENCY:}
\begin{itemize}[leftmargin=*, itemsep=1pt, topsep=2pt]
\item Fire damage must be uneven and spatially coherent, with heavier damage near the likely fire origin and fading outward. 
\item Upper surfaces, ceilings, and the tops of walls show heavier soot accumulation and heat damage than lower surfaces (hot gas layer rises). 
\item Visible soot gradients and runoff streaking on vertical surfaces. 
\item V-patterns and inverted-cone burn patterns on walls indicate upward fire spread from point sources. 
\item Floors show debris accumulation from collapsed shelving, ceiling, and wall-mounted objects. 
\item Adjacent rooms or hallways visible through doorways should show diminishing damage with distance from the fire origin. 
\end{itemize}

\textbf{SECONDARY EFFECTS:} 
\begin{itemize}[leftmargin=*, itemsep=1pt, topsep=2pt] 
\item Realistic firefighting residue including white or gray extinguisher powder and water pooling or streaking on floors. 
\item Some surfaces may appear damp, washed, or partially streaked from hose water. 
\item Debris, ash, and fragmented material accumulate naturally around collapse points, along walls, and at the base of furniture. 
\item Electrical wiring may be exposed where wall or ceiling coverings have burned away. 
\item Smoke detectors, light fixtures, and ceiling-mounted objects may hang partially detached. 
\end{itemize}

\textbf{OBJECT DISTRIBUTION:} 
\begin{itemize}[leftmargin=*, itemsep=1pt, topsep=2pt] 
\item Scene should feel asymmetric, naturally disturbed, and realistic. 
\item Include a mix of partially recognizable objects and heavily degraded objects. 
\item No central hero object and no staged or overly clean composition. 
\item Natural occlusion and layering should occur (objects partially buried under debris, leaning against walls, fallen from shelves). 
\item At least 2-3 objects should be identifiable to a trained observer despite damage. 
\end{itemize}

\textbf{STRUCTURAL DETAILS:} 
\begin{itemize}[leftmargin=*, itemsep=1pt, topsep=2pt] 
\item Possible partial collapse of ceilings, shelving, partition walls, door frames, or built-in cabinetry. 
\item Exposed structural elements (studs, joists, wiring, plumbing) where wall and ceiling coverings have failed. 
\item Doors may be open, partially burned, or missing. Door hardware (knobs, hinges) typically survives. 
\item Windows may be broken out by thermal shock or firefighting ventilation. Curtain rods may remain while curtain fabric is gone. 
\end{itemize}

\textbf{CAMERA STYLE:} 
Realistic handheld or inspection-style photograph, approximately 24-35mm equivalent, natural framing imperfections, documentary-like realism. No cinematic grading or artistic stylization. 

\textbf{STRICT CONSTRAINTS:} 
No flames, no active smoke or haze effects, no stylization, no painterly or artistic rendering, no cinematic grading, no unrealistic symmetry, no pristine clean areas unless physically justified (e.g., shielded by another object), no duplicated obvious objects or synthetic visual artifacts.

\texttt{\{scene\_specific\}}

The output must be indistinguishable from a real-world post-fire documentation photo.
\end{tcolorbox}

To reduce visual homogeneity, scene prompts vary scene type, damage level, material profile, clutter, lighting, socioeconomic style, and layout, and explicitly prohibit copying the reference layout or viewpoint.

\subsection{Grounding Judge and Filtering}
\label{app:gen-filtering}

After generation, each scene candidate is scored by a Gemini~3 Flash vision-language judge that compares the candidate against its real-image references on a 0-1 grounding scale.

\begin{tcolorbox}[colback=gray!5, colframe=gray!85, breakable, title={\small\textbf{Grounding Judge Prompt}}] 
\small 
You are a grounding judge. Compare the candidate post-fire scene image against its reference real-world images. Score grounding on a 0 to 1 scale based on: (1) realism, (2) material and fire-physics fidelity, (3) structural damage plausibility, and (4) overall consistency with real-world post-fire visual characteristics while preserving novelty. 

Return strict JSON only:\\ 
\texttt{\{"grounding\_score": <float>, "reasoning": "<short>", "pass": <bool>\}} Set pass = true only if grounding\_score $\geq$ 0.7. 

\end{tcolorbox} 

Candidates scoring below 0.7 are rejected. This filtering yields 21{,}400 accepted scenes from approximately 29{,}000 generated candidates, for an acceptance rate of 73.8\%. The grounding score is not used as an evaluation label. It is used only to filter generated scenes for realism, material and fire-physics fidelity, structural damage plausibility, and consistency with the real post-fire reference set while preserving novelty. Object boxes and labels are then handled as annotations for the detection task rather than as outputs of the grounding judge.

Initial object annotations are produced automatically and then verified or corrected during dataset quality control. For annotation reliability, two authors independently assigned accept/reject labels on a stratified 300-object subset, yielding 93.4\% agreement and Cohen's $\kappa=0.836$.

Initial bounding boxes are produced automatically and then visually verified or corrected by the authors before evaluation.

\section{Evaluation Task Details}
\label{app:task-details}

This appendix specifies the five evaluation tasks from Sec.~\ref{sec:benchmark}, namely degraded-object detection, pristine-state recovery and retrieval, original material recovery, pristine description generation, and functional reasoning.

\subsection{Degraded-Object Detection}
\label{app:detection-task}

Given a scene-level post-fire image, the model detects and localizes objects of interest. This task evaluates whether a detector can recover object remnants in cluttered, physically degraded scenes.

We evaluate RF-DETR outputs using class-aware COCO-mapped mAP@0.5:0.95, reported overall and separately by degradation level. A manually constructed lookup table maps \benchmark object classes to the nearest COCO category for compatibility with COCO-pretrained detectors. Only boxes with unambiguous COCO mappings are included in mAP computation. The mapping covers 55 of the 80 COCO categories. The 55 mapped COCO categories cover the most common indoor object types (furniture, electronics, containers, appliances). Excluded categories are primarily \benchmark-specific items without clear COCO counterparts (e.g., aerosol cans, gas cylinders, specific tool subtypes). The full mapping table is released with the evaluation code. A prediction is counted as correct only when it satisfies both localization and mapped-category correctness. Detection is evaluated on the scene-level subset and no natural-language prompt is used.

\subsection{Pristine-State Recovery and Retrieval}
\label{app:pristine-retrieval}

Given a degraded object image, the goal is to recover its paired pristine state. We evaluate this task with two complementary protocols: feature-space alignment and gallery retrieval.

\boldheader{Feature-space alignment.}
For frozen encoder diagnostics, CLIP ViT-L/14 and SigLIP2 features are recovered toward paired pristine targets. Given degraded features $z^\ell$ and paired pristine features $z^p$, \module produces recovered features $\tilde{z}^\ell$. We report mean squared error (MSE) and mean token cosine similarity between $\tilde{z}^\ell$ and $z^p$. Lower MSE and higher cosine similarity indicate stronger pristine alignment.

\boldheader{Gallery retrieval.}
For VLM retrieval, the model retrieves the corresponding pristine reference from a gallery of pristine objects. For consistency with the VLM setting, we use the following natural-language query:

\begin{quote}\small\ttfamily
Find the pristine object image in the gallery that corresponds to this physically degraded object.
\end{quote}

The prompt defines the retrieval intent, but scoring is performed in feature space: the model's visual encoder extracts features for the degraded query and all pristine gallery images, and gallery images are ranked by cosine similarity. Each degraded query has exactly one paired pristine target. We report Recall@1 as the main gallery-retrieval metric.

\subsection{Original Material Recovery}
\label{app:material-recovery}

Given a degraded object image, the model predicts the object's pre-degradation materials. The prompt is:

\begin{quote}\small\ttfamily
What were the main materials of this object before degradation? Reply with a comma-separated list of single-word materials only (for example: wood, steel, plastic). Do not describe the damage.
\end{quote}

Ground-truth materials are parsed from the object's part-level material annotations and mapped to the 11 normalized material classes in \benchmark. Model responses are lowercased, stripped of punctuation, singularized, and mapped to the same material vocabulary using a fixed synonym table. Let $Y_i^{\mathrm{mat}}$ and $\hat{Y}_i^{\mathrm{mat}}$ denote the ground-truth and predicted material sets for object $i$. We compute micro-F1 across all objects and material labels:
\[
P_{\mathrm{mat}} =
\frac{\sum_i |\hat{Y}_i^{\mathrm{mat}} \cap Y_i^{\mathrm{mat}}|}
{\sum_i |\hat{Y}_i^{\mathrm{mat}}|},
\qquad
R_{\mathrm{mat}} =
\frac{\sum_i |\hat{Y}_i^{\mathrm{mat}} \cap Y_i^{\mathrm{mat}}|}
{\sum_i |Y_i^{\mathrm{mat}}|},
\]
\[
\mathrm{F1}_{\mathrm{mat}} =
\frac{2P_{\mathrm{mat}}R_{\mathrm{mat}}}
{P_{\mathrm{mat}} + R_{\mathrm{mat}}}.
\]
If a model predicts no labels for a task, precision is set to zero; if the ground-truth set is empty, recall is set to zero. In our benchmark, ground-truth material and function sets are non-empty. We report $100 \cdot \mathrm{F1}_{\mathrm{mat}}$.

\subsection{Pristine Description Generation}
\label{app:pristine-description}

Given a degraded object image, the model generates a one-sentence description of the object's pre-degradation state. The prompt is:

\begin{quote}\small\ttfamily
Describe what this object most likely was before it was damaged. Mention its object type, main material(s), and intended use. Do not focus on the damage. Answer in one concise sentence.
\end{quote}

Ground-truth identity is taken from the object manifest, materials from the part-level annotations, and function from the canonical function tag assigned to the object category. Each generated sentence is parsed into three fields: object identity, materials, and function. Identity and function predictions are lowercased, stripped of punctuation and articles, singularized, and matched to the canonical category or function vocabulary using fixed synonym tables. Materials are normalized as in Appendix~\ref{app:material-recovery}.

For object $i$, identity is scored as
\[
S_i^{\mathrm{id}} =
\mathbb{1}[\hat{y}_i^{\mathrm{id}} = y_i^{\mathrm{id}}],
\]
where $y_i^{\mathrm{id}}$ is the ground-truth object category and $\hat{y}_i^{\mathrm{id}}$ is the normalized predicted category. Function is scored analogously:
\[
S_i^{\mathrm{func}} =
\mathbb{1}[\hat{y}_i^{\mathrm{func}} = y_i^{\mathrm{func}}].
\]
Materials are scored using micro-F1 over the normalized material vocabulary, following Appendix~\ref{app:material-recovery}. The final pristine-description score is the unweighted mean of identity accuracy, material micro-F1, and function accuracy:
\[
S_{\mathrm{desc}} =
100 \cdot \frac{1}{3}
\left(
\frac{1}{N}\sum_{i=1}^{N} S_i^{\mathrm{id}}
+
\mathrm{F1}_{\mathrm{mat}}
+
\frac{1}{N}\sum_{i=1}^{N} S_i^{\mathrm{func}}
\right).
\]

When deterministic synonym normalization cannot resolve a paraphrase, we use a deterministic LLM equivalence judge as a fallback. The judge receives only the predicted phrase, the ground-truth label, and the relevant synonym policy, and returns a binary equivalent/not-equivalent decision. It is run with temperature $0.0$ and is used only for unresolved identity and function paraphrases, not for material scoring.

The synonym table resolves $5{,}562/6{,}250$ identity predictions ($89.0\%$) and $5{,}312/6{,}250$ function predictions ($85.0\%$) without invoking the judge. The GPT-5 fallback is triggered for $688/6{,}250$ identity evaluations ($11.0\%$) and $938/6{,}250$ function evaluations ($15.0\%$), mainly for paraphrases such as ``writing surface'' for ``desk.'' Replacing the fallback with a strict no-match rule changes the aggregate description score by less than $0.8$ points.

\subsection{Functional Reasoning}
\label{app:functional-task}

Given a degraded object image, the model answers two structured questions about the object's pre-degradation function and restoration needs without being told its category.

\boldheader{Prompt.}

\begin{quote}\small\ttfamily
You are looking at an object that has been damaged in a fire. Reason about what the object originally was and answer two questions about its pre-damage identity and restoration potential.

\medskip FUNCTION: In one short phrase (5 words or fewer), what was this object's original function or primary use?

\medskip REPLACEMENTS: List the main materials that would be needed to restore this object to working condition. Reply as a comma-separated list of single-word items.

\medskip Respond in this exact format with no extra text:\\
FUNCTION: <your answer>\\
REPLACEMENTS: <comma-separated list>
\end{quote}

The prompt is identical across all hosts and across \module-on/off conditions. All calls use temperature $0.0$ and a maximum of 150 output tokens.

\boldheader{Ground truth.}
\textbf{i. Function labels.}
We define 20 canonical function tags, such as seating, lighting, cooking\_appliance, food\_storage, general\_storage, safety\_equipment, and communication\_device. Each object type in \benchmark is mapped to one or two function tags. The mapping is constructed with LLM assistance and verified by the authors. The full mapping is released as supplementary material.

\textbf{ii. Restoration-material labels.}
The canonical material set annotated for each object type serves as the ground-truth restoration-material set. This reuses existing material annotations and introduces no additional task-specific labels.

\boldheader{Scoring.}
The FUNCTION response is normalized by lowercasing, stripping punctuation, singularizing, and mapping to the fixed function-tag vocabulary using keyword and synonym rules. When this deterministic mapping is ambiguous, we use the same deterministic LLM fallback described in Appendix~\ref{app:pristine-description}. The judge receives the model response and the fixed function-tag vocabulary, and returns one or more canonical tags. It is not given the ground-truth tag. Let $Y_i^{\mathrm{func}}$ and $\hat{Y}_i^{\mathrm{func}}$ denote the ground-truth and predicted function-tag sets for object $i$. Function performance is computed as micro-F1:
\[
P_{\mathrm{func}} =
\frac{\sum_i |\hat{Y}_i^{\mathrm{func}} \cap Y_i^{\mathrm{func}}|}
{\sum_i |\hat{Y}_i^{\mathrm{func}}|},
\qquad
R_{\mathrm{func}} =
\frac{\sum_i |\hat{Y}_i^{\mathrm{func}} \cap Y_i^{\mathrm{func}}|}
{\sum_i |Y_i^{\mathrm{func}}|},
\]
\[
\mathrm{F1}_{\mathrm{func}} =
\frac{2P_{\mathrm{func}}R_{\mathrm{func}}}
{P_{\mathrm{func}} + R_{\mathrm{func}}}.
\]

The REPLACEMENTS response is parsed as a comma-separated material set and normalized to the same 11 material classes used for material recovery. Let $Y_i^{\mathrm{rest}}$ and $\hat{Y}_i^{\mathrm{rest}}$ denote the ground-truth and predicted restoration-material sets. Restoration-material performance is computed as micro-F1:
\[
P_{\mathrm{rest}} =
\frac{\sum_i |\hat{Y}_i^{\mathrm{rest}} \cap Y_i^{\mathrm{rest}}|}
{\sum_i |\hat{Y}_i^{\mathrm{rest}}|},
\qquad
R_{\mathrm{rest}} =
\frac{\sum_i |\hat{Y}_i^{\mathrm{rest}} \cap Y_i^{\mathrm{rest}}|}
{\sum_i |Y_i^{\mathrm{rest}}|},
\]
\[
\mathrm{F1}_{\mathrm{rest}} =
\frac{2P_{\mathrm{rest}}R_{\mathrm{rest}}}
{P_{\mathrm{rest}} + R_{\mathrm{rest}}}.
\]

The final functional-reasoning score is the unweighted mean of function-tag micro-F1 and restoration-material micro-F1:
\[
S_{\mathrm{func}} =
100 \cdot \frac{1}{2}
\left(
\mathrm{F1}_{\mathrm{func}}
+
\mathrm{F1}_{\mathrm{rest}}
\right).
\]
If the denominator of precision or recall is zero, the corresponding value is set to zero.

\boldheader{Baseline artifact.}
At high severity, models sometimes default to coarse answers such as ``container'' or ``furniture.'' Such responses may partially match broad function tags even when the predicted object identity is underspecified. We therefore interpret functional-reasoning results together with pristine-description and retrieval performance. \module's high-severity gains, when accompanied by improved identity and material recovery, indicate more specific pre-degradation recovery rather than only coarse functional guessing.

\section{\module Implementation Details}
\label{app:frm-hosts}

This appendix details the host-specific tap points used to instantiate \module. All hosts follow the same formulation from Sec.~\ref{sec:frm}: degraded features and paired pristine targets are extracted at the same frozen feature interface, \module maps $z^\ell\in\mathbb{R}^{B\times N\times D}$ to $\tilde{z}^{\ell}$, and only \module is trained. The token dimension $D$ is determined by the selected tap point, while $N$ is the number of visual tokens at that interface.

\begin{table}[t]
\centering\small
\setlength{\tabcolsep}{4pt}
\caption{\textbf{\module parameter overhead across VLM and frozen-encoder hosts.}
For each host, $D$ is the hidden dimension at the selected tap point. All listed hosts use four Transformer blocks with MLP ratio $r{=}4$, giving $P_{\module}\approx M(4+2r)D^2$. Overhead is reported relative to the host parameter count where the host size is fixed in our experiments.}
\begin{tabular}{lcccccc}
\toprule
Host & Tap & $D$ & $M$ & $r$ & $P_{\module}$ & Overhead \\
\midrule
Qwen3-VL-4B & post-merger & 2560 & 4 & 4 & 314.57M & 7.86\% \\
Qwen2.5-VL-3B & post-merger & 2048 & 4 & 4 & 201.33M & 6.71\% \\
InternVL3.5-4B & post-projector & 2560 & 4 & 4 & 314.57M & 7.86\% \\
Molmo2-4B & post-projector & 2560 & 4 & 4 & 314.57M & 7.86\% \\
LLaVA-OV-1.5-4B & post-merger & 2560 & 4 & 4 & 314.57M & 7.86\% \\
CLIP ViT-L/14 & pre-projection & 1024 & 4 & 4 & 50.33M & -- \\
SigLIP2-base & pre-projection & 768 & 4 & 4 & 28.31M & -- \\
\bottomrule
\end{tabular}
\label{tab:frm-overhead}
\end{table}

Table~\ref{tab:frm-overhead} reports the resulting \module sizes for VLM and frozen-encoder hosts. Qwen3-VL, InternVL3.5, Molmo2, and LLaVA-OV-1.5 use $D{=}2560$ at their LLM-facing post-merger or post-projector interfaces, yielding $314.57$M \module parameters. CLIP and SigLIP2 use earlier pre-projection vision tokens, giving $50.33$M and $28.31$M \module parameters respectively. We omit percentage overhead for CLIP and SigLIP2 because the relevant denominator depends on whether one counts only the vision tower or the full contrastive model. 

\begin{figure}[t]
    \centering
    \includegraphics[width=0.65\linewidth]{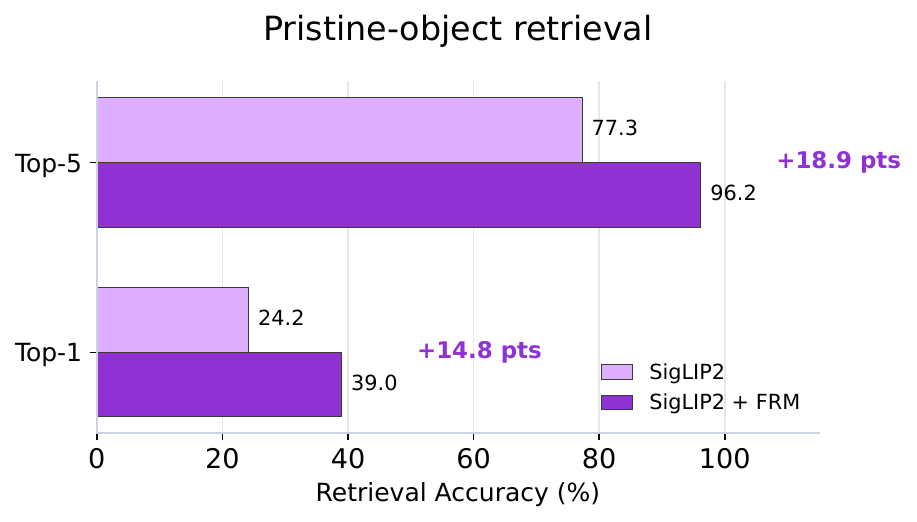}
    \caption{\textbf{SigLIP2 pristine-state retrieval.} Given a degraded object query, SigLIP2 ranks candidate pristine gallery images by feature similarity. \module improves degraded-to-pristine retrieval, increasing Top-1 from $24.2\%$ to $39.0\%$ and Top-5 from $73.3\%$ to $96.2\%$, approaching the pristine-query ceiling of $100\%$.}
    \label{fig:siglip-retrieval} 
\end{figure}

For SigLIP2 retrieval, the query is the degraded object image and the gallery contains candidate pristine object images (Figure~\ref{fig:siglip-retrieval}).

\subsection{\module Instantiation in VLM Hosts}
\label{app:frm-vlm-hosts}

\boldheader{InternVL3.5-4B.}
For InternVL3.5, \module is trained at the post-projector visual-token interface. The frozen InternVL visual encoder first produces image tokens and the frozen multimodal projector maps them into the language-model input space. We apply \module after this projector, so both degraded features and paired pristine targets have shape $z\in\mathbb{R}^{B\times N\times2560}$. With $448\times448$ inputs, the token count $N$ is determined by the model preprocessing and projector configuration, and the feature dimension is the text hidden size $D{=}2560$. During training, pristine post-projector features are cached once, degraded images are encoded on the fly, and \module is optimized with the same MSE plus cosine recovery loss used in Sec.~\ref{sec:frm}. At inference, the InternVL host remains frozen and the wrapper replaces the live degraded post-projector tokens with \module-corrected tokens before generation. The baseline and \module conditions therefore differ only in the visual feature path consumed by the frozen language model.

\boldheader{LLaVA-OneVision-1.5-4B.}
For LLaVA-OneVision-1.5, \module is trained at the post-merger visual-token interface. The frozen Rice ViT first produces patch tokens, and the frozen RicePatchMerger spatially downsamples and projects them into the LLM-facing hidden space. We insert \module after this merger, so the tapped tensor has shape $z\in\mathbb{R}^{B\times256\times2560}$ for the standard $448\times448$ input setting. Here $N{=}256$ follows from $(448/14/2)^2$, and $D{=}2560$ is the merger output dimension. The implementation reads the exact token count and feature dimension from the model configuration. Pristine post-merger tokens are cached before training, degraded tokens are encoded on the fly, and \module is trained with the same paired feature objective as in Sec.~\ref{sec:frm}. At inference, the wrapper patches the visual forward pass only inside the \module-enabled context, applies \module independently to each packed image token sequence, and then restores the original path. Thus, LLaVA generation, prompting, processor behavior, and language-model weights remain unchanged.

\boldheader{Molmo2-4B.}
For Molmo2, \module is trained at the post-projector visual-token interface used by the host before language generation. The frozen vision stack produces packed visual tokens, the frozen projector maps them into the LLM-facing feature space, and \module is applied to those post-projector tokens. Thus, the tapped tensor has shape $z\in\mathbb{R}^{B\times N\times2560}$, where $N$ is the post-projector visual-token count and $D{=}2560$ is the LLM-facing hidden dimension read from the host configuration. We use the same VLM \module configuration with $M{=}4$, $8$ attention heads, and MLP ratio $r{=}4$.

Training follows the paired feature-recovery setup in Sec.~\ref{sec:frm}. Pristine post-projector features are cached once, degraded images are encoded on the fly, and \module is optimized with the same MSE plus cosine alignment loss against the paired pristine target. At inference, the wrapper enables \module only inside the \module-on condition. It runs the frozen visual path, splits packed visual features by image using runtime grid metadata, applies \module independently to each image token sequence, and concatenates the corrected features before language generation. Baseline, \module, and pristine-ceiling evaluations therefore share the same prompts, decoding, and frozen language model, differing only in whether degraded post-projector visual tokens are corrected before being consumed by the model.

\subsection{\module Instantiation in Frozen Encoders}
\label{app:frm-frozen-encoders}

\boldheader{CLIP ViT-L/14.}
For CLIP, \module is trained on pre-projection visual tokens from the frozen CLIP vision tower. We tap the CLIPVisionModel last hidden state before the visual projection into the text-aligned embedding space. For CLIP only, we drop the CLS token so that \module operates on spatial patch tokens. With $224\times224$ inputs and patch size $14$, the tapped feature tensor has $N{=}256$ patch tokens and hidden dimension $D{=}1024$, giving $z\in\mathbb{R}^{B\times256\times1024}$. Pristine patch-token features are cached once, degraded images are encoded on the fly, and \module predicts recovered tokens in the same $B\times N\times D$ space. The frozen CLIP encoder is never updated, and evaluation compares recovered degraded tokens to paired pristine CLIP tokens using MSE, cosine similarity, and pooled-token retrieval.

\boldheader{SigLIP2.}
For SigLIP2, \module is also trained at the pre-projection visual-token interface. We tap the frozen SigLIP2 vision model's last hidden state and use patch tokens rather than pooled image embeddings or text-aligned outputs. With google/siglip2-base-patch16-224, the input size is $224\times224$, the patch size is $16$, and the visual grid is $14\times14$. The tapped tensor therefore has $N{=}196$ patch tokens and hidden dimension $D{=}768$, giving $z\in\mathbb{R}^{B\times196\times768}$. As in Sec.~\ref{sec:frm}, \module is shape preserving: degraded features $z^\ell$ are mapped to recovered features $\tilde{z}^\ell$ and supervised against cached paired pristine features $z^p$ from the same tap. We keep the SigLIP2 encoder frozen and report feature-space recovery metrics as well as pristine-state retrieval using pooled visual features.

\paragraph{Integration and tap point.}
Correcting Qwen3-VL's DeepStack stream slightly reduces identity accuracy, so we keep \module at the primary visual-token interface. For tap-point selection, downstream-facing taps outperform earlier visual-token taps. Qwen3-VL improves from 72.90 R@1 with a pre-merger tap to 85.28 with a post-merger tap, while InternVL3.5 improves from 63.84 R@1 with a pre-projector tap to 74.92 with a post-projector tap (Appendix~\ref{tab:app-frm-ablations}). These results support correcting the features consumed by the frozen language model.

\section{Per-Level VLM Results}
\label{app:vlm-per-level}

\begin{figure}[ht]
    \centering
    \includegraphics[width=\linewidth]{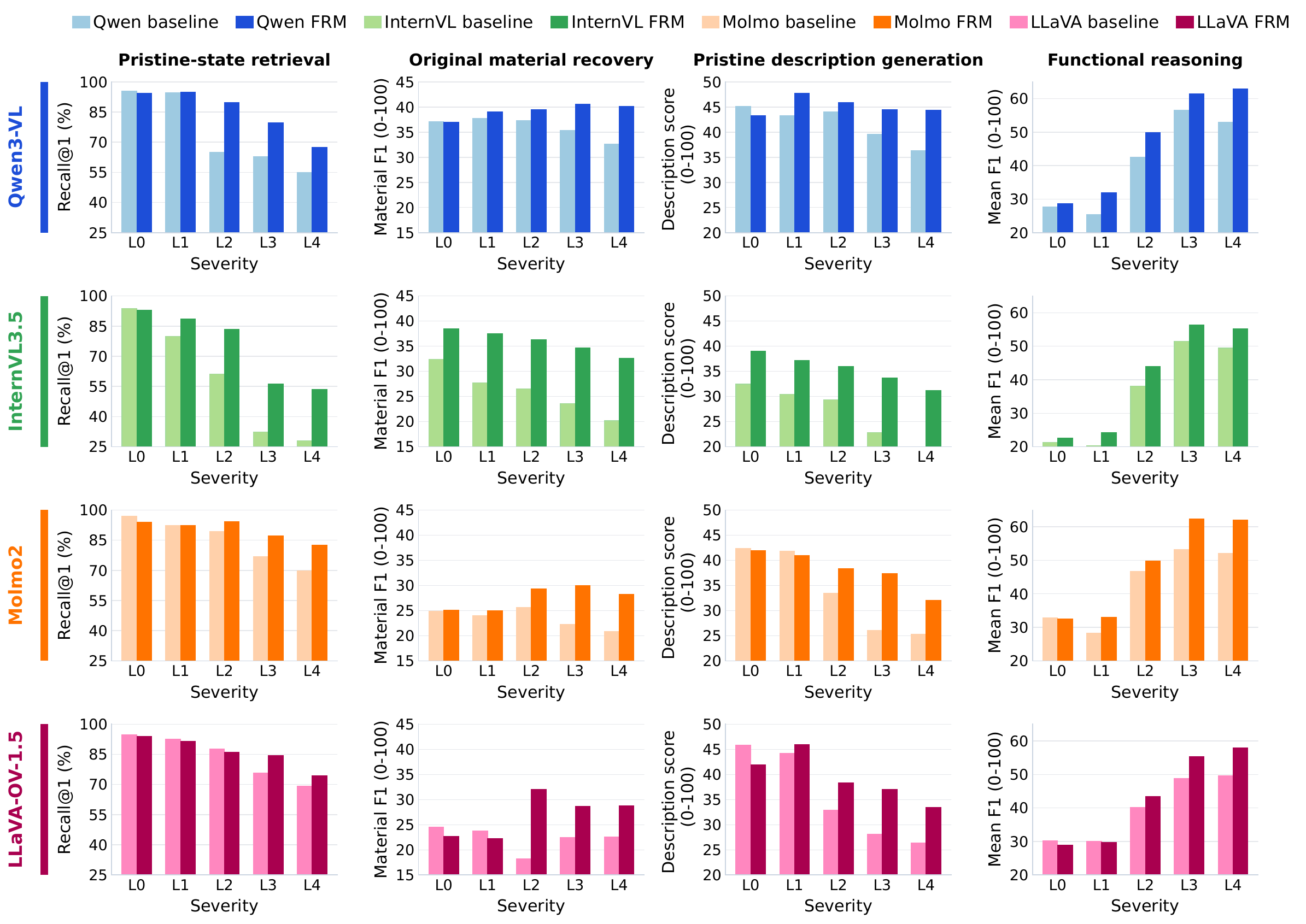}
    \caption{\textbf{Per-level VLM pristine-state recovery across additional hosts.} Baseline models are shown in light colors and \module-enabled models in dark colors across L0-L4. Columns report pristine-state retrieval (R@1), original material recovery (F1), pristine description generation (score), and functional reasoning (mean F1). \module gains concentrate at higher severity across hosts.}
    \label{fig:all-vlm-severity}
    \vspace{-1.2em}
\end{figure}

Figure~\ref{fig:all-vlm-severity} and table~\ref{tab:vlm-per-level-full} provide the full per-level breakdown of \module gains across all five frozen VLM hosts and four object-level pristine-state tasks. Gains are small or occasionally negative at L0-L1, where degradation is mild and baseline features are already close to the pristine state. Gains concentrate at L2-L4, where physical transformation causes larger feature displacement and the learned correction becomes more useful. This pattern supports the view that \module is most beneficial when degradation induces structured feature drift.

\begin{table*}[ht]
\centering
\setlength{\tabcolsep}{3.2pt}
\caption{\textbf{Per-level VLM pristine-state recovery.}
All scores are on a 0-100 scale. Mean Base and Mean +\module report the average score over L0-L4. Mean $\Delta$ is the average improvement, while $\Delta$L0-$\Delta$L4 report per-level changes from adding \module to the frozen host. Gains are strongest at L2-L4, while L0-L1 can show smaller gains, matched performance, or occasional drops.}
\renewcommand{\arraystretch}{0.95}
\rowcolors{3}{gray!8}{white}
\begin{tabular}{lrrrrrrrr}
\toprule
\rowcolor{gray!22}
Model & Mean Base & Mean +\module & Mean $\Delta$ & $\Delta$L0 & $\Delta$L1 & $\Delta$L2 & $\Delta$L3 & $\Delta$L4 \\
\midrule

\rowcolor{gray!18}
\multicolumn{9}{l}{\textbf{Pristine-state retrieval (R@1)}} \\
Qwen3-VL       & 74.55 & \textbf{85.28} & $+10.73$ & $-1.06$ & $+0.36$ & $+24.88$ & $+16.88$ & $+12.58$ \\
Qwen2.5-VL     & 60.20 & \textbf{72.03} & $+11.83$ & $+0.90$ & $+4.36$ & $+17.33$ & $+19.78$ & $+16.78$ \\
InternVL3.5    & 59.09 & \textbf{74.92} & $+15.83$ & $-0.83$ & $+8.46$ & $+22.18$ & $+23.84$ & $+25.48$ \\
Molmo2         & 85.06 & \textbf{90.00} & $+4.94$  & $-3.07$ & $+0.00$ & $+5.00$  & $+10.23$ & $+12.53$ \\
LLaVA-OV-1.5   & 83.97 & \textbf{86.09} & $+2.12$  & $-0.77$ & $-1.09$ & $-1.54$  & $+8.85$  & $+5.15$  \\
\midrule

\rowcolor{gray!18}
\multicolumn{9}{l}{\textbf{Original material recovery (F1)}} \\
Qwen3-VL       & 36.00 & \textbf{39.24} & $+3.24$ & $-0.13$ & $+1.39$ & $+2.19$  & $+5.22$  & $+7.53$  \\
Qwen2.5-VL     & 30.50 & \textbf{36.78} & $+6.28$ & $+0.93$ & $+4.29$ & $+6.50$  & $+8.50$  & $+10.90$ \\
InternVL3.5    & 26.10 & \textbf{35.91} & $+9.81$ & $+6.06$ & $+9.70$ & $+9.79$  & $+11.14$ & $+12.39$ \\
Molmo2         & 23.56 & \textbf{27.52} & $+3.96$ & $+0.20$ & $+0.93$ & $+3.66$  & $+7.64$  & $+7.37$  \\
LLaVA-OV-1.5   & 22.31 & \textbf{26.89} & $+4.58$ & $-1.77$ & $-1.54$ & $+13.85$ & $+6.15$  & $+6.20$  \\
\midrule

\rowcolor{gray!18}
\multicolumn{9}{l}{\textbf{Pristine description generation (score)}} \\
Qwen3-VL       & 41.67 & \textbf{45.15} & $+3.48$ & $-1.83$ & $+4.40$ & $+1.86$ & $+4.95$  & $+8.01$  \\
Qwen2.5-VL     & 40.50 & \textbf{44.00} & $+3.50$ & $+1.43$ & $+3.45$ & $+2.00$ & $+4.87$  & $+6.90$  \\
InternVL3.5    & 27.01 & \textbf{35.42} & $+8.41$ & $+6.55$ & $+6.73$ & $+6.64$ & $+10.83$ & $+11.32$ \\
Molmo2         & 33.80 & \textbf{38.13} & $+4.33$ & $-0.37$ & $-0.87$ & $+4.91$ & $+11.24$ & $+6.72$  \\
LLaVA-OV-1.5   & 35.48 & \textbf{39.34} & $+3.86$ & $-3.83$ & $+1.70$ & $+5.41$ & $+8.93$  & $+7.09$  \\
\midrule

\rowcolor{gray!18}
\multicolumn{9}{l}{\textbf{Functional reasoning (mean F1)}} \\
Qwen3-VL       & 41.02 & \textbf{46.96} & $+5.94$ & $+0.96$ & $+6.59$ & $+7.30$ & $+4.88$ & $+9.93$  \\
Qwen2.5-VL     & 38.11 & \textbf{43.62} & $+5.51$ & $+1.12$ & $+5.60$ & $+4.68$ & $+7.70$ & $+8.45$  \\
InternVL3.5    & 36.13 & \textbf{40.48} & $+4.35$ & $+1.30$ & $+3.92$ & $+5.78$ & $+4.95$ & $+5.80$  \\
Molmo2         & 42.64 & \textbf{47.95} & $+5.31$ & $-0.36$ & $+4.67$ & $+3.12$ & $+9.11$ & $+10.01$ \\
LLaVA-OV-1.5   & 39.76 & \textbf{43.07} & $+3.31$ & $-1.40$ & $+0.29$ & $+3.33$ & $+6.60$ & $+8.27$  \\
\bottomrule
\end{tabular}
\label{tab:vlm-per-level-full}
\end{table*}

\section{Frozen-Encoder Feature Recovery}
\label{app:frozen-encoder-recovery}

We evaluate \module on frozen CLIP ViT-L/14 and SigLIP2 encoders to isolate feature-space recovery from detector and language-model effects. The input is a degraded object image, and the target is the paired pristine image of the same object. We compare recovered degraded features against paired pristine features at the encoder tap point.

For retrieval, the query is the degraded object image and the gallery contains candidate pristine object images. SigLIP2 ranks gallery images by visual-feature similarity, with one paired pristine target per query. The pristine-query ceiling is $100\%$ because querying with the pristine image retrieves the identical reference from the gallery.

\section{Additional \module Ablations}
\label{app:frm-ablations}

\paragraph{DeepStack and tap-point selection.}
Table~\ref{tab:app-frm-ablations} reports additional integration ablations. For Qwen3-VL, applying \module to the DeepStack stream slightly reduces identity accuracy compared with correcting only the primary visual-token interface. For tap-point selection, later LLM-facing interfaces outperform earlier visual-token interfaces. Qwen3-VL improves from 72.90 R@1 with a pre-merger tap to 85.28 with a post-merger tap, and InternVL3.5 improves from 63.84 R@1 with a pre-projector tap to 74.92 with a post-projector tap.

\begin{table}[t]
\centering
\footnotesize
\caption{\textbf{Additional \module ablations.}
DeepStack correction does not improve Qwen3-VL identity accuracy. Downstream-facing taps, post-merger for Qwen3-VL and post-projector for InternVL3.5, outperform earlier visual-token taps for retrieval.}
\setlength{\tabcolsep}{5pt}
\begin{tabular}{lc}
\toprule
Configuration & Score \\
\midrule
\multicolumn{2}{l}{\textbf{DeepStack, Qwen3-VL, ID acc.$\uparrow$}} \\
No \module                    & 32.69 \\
\module, DeepStack off        & \textbf{37.31} \\
\module, DeepStack on         & 36.54 \\
Pristine ceiling              & 43.08 \\
\midrule
\multicolumn{2}{l}{\textbf{Tap point, retrieval R@1$\uparrow$}} \\
Qwen3-VL post-merger          & \textbf{85.28} \\
Qwen3-VL pre-merger           & 72.90 \\
InternVL3.5 post-projector    & \textbf{74.92} \\
InternVL3.5 pre-projector     & 63.84 \\
\bottomrule
\end{tabular}
\label{tab:app-frm-ablations}
\end{table}




\newpage

\end{document}